\documentclass[11pt]{article}

\usepackage[margin=1in]{geometry}
\usepackage{times}
\usepackage{microtype}

\usepackage{amsmath,amssymb,amsthm}
\usepackage{mathtools}
\usepackage{bm}

\usepackage{graphicx}
\usepackage{booktabs}
\usepackage{multirow}
\usepackage{array}
\usepackage{subfig}
\usepackage{tabularx}
\usepackage{amsfonts}
\usepackage{amsmath}
\usepackage{amsthm}
\usepackage{amssymb}
\usepackage{enumerate}
\usepackage{xcolor}
\usepackage{url}
\usepackage{hyperref}
\hypersetup{
    colorlinks=true,
    linkcolor=blue,
    citecolor=blue,
    urlcolor=blue
}

\usepackage[numbers,sort&compress]{natbib}

\theoremstyle{definition}
\newtheorem{definition}{Definition}

\theoremstyle{remark}

\title{Explicit Trajectory Diversity for RL-Based Post-Training of LLM Agents}

\author{
    Huaiyu Fu$^{1}$ \qquad
    Heng Cao$^{2}$ \qquad
    Hao Wang \qquad
    Jian Yao \qquad
    Tao Chen
    \\[3pt]
    $^{1}$Microsoft\\
    $^{2}$tuyoogame\\
}

\begin{document}

\maketitle

\begin{abstract}
Large language model (LLM) agents often admit multiple high-quality solutions to the same task, differing in reasoning structure, tool-use pattern, or interaction trajectory.
Yet existing notions of diversity in LLM post-training are mostly implicit, arising from general stochasticity and regularization mechanisms rather than explicitly targeting task-relevant behavioral variation.
While such implicit diversity can be useful, it does not directly specify which forms of behavioral variation should be encouraged for a given task.
In this work, we study \emph{explicit trajectory diversity} in RL-based post-training for LLMs.
Our key idea is to define diversity through user-specified, task-specific trajectory descriptors, which map each sampled trajectory to an interpretable behavioral representation, and then measure diversity as a set-level functional over the resulting descriptor matrix.
Building on this formulation, we introduce \emph{Trajectory-guided Joint Policy Optimization} (TJPO), a single-policy framework that optimizes explicit diversity over sampled trajectory groups, avoiding the need for population-based policy training, and instantiate it within group-based policy optimization through trajectory-level learning signals.
This design makes the diversity objective both interpretable and controllable: practitioners can specify whether diversity should correspond to action composition, location visitation patterns, spatial coverage, or other task-relevant behavioral dimensions.
Experiments on Sokoban and ALFWorld show that TJPO improves task-specific trajectory diversity while maintaining competitive task performance.
Descriptor and trajectory analyses show that the learned variation follows the specified behavioral dimensions and includes distinct successful strategies.
Beyond diversity metrics, evaluations on an ALFWorld subset with verified alternative solutions demonstrate practical benefits without additional training: compared with GiGPO, TJPO improves compliant success under user-specified location constraints by 8.3 percentage points and task success under environmental disruptions by 7.5 percentage points.
These findings suggest that explicitly shaping trajectory diversity can help LLM agents satisfy user requirements and remain effective when task conditions change.
\end{abstract}

\section{Introduction}
Large language models (LLMs) are increasingly deployed as agents that solve multi-step tasks through planning, interaction, and tool use~\citep{du2026asurvey,zhang2026thelandscape}.
In such settings, a single objective often admits multiple high-quality solutions that differ substantially in their behavioral patterns, such as plan decomposition, tool selection, and interaction trajectories~\citep{yao2022webshop,shridhar2020alfworld,schrader2018gymsokoban}.
For example, consider a household agent asked to place a target object in a desired location.
One successful strategy may first explore the environment to identify relevant objects and receptacles, verify object states, and then execute the manipulation sequence.
Another strategy may directly exploit known task regularities, navigate to likely object locations, and perform a shorter sequence of actions.
Both strategies can solve the task, but they differ in object-interaction patterns, action composition, and robustness to changes in the environment.
Similarly, in a planning game such as Sokoban, two successful solutions may reach the same goal state while following different routes, pushing boxes in different orders, or using different intermediate subgoals.
Therefore, beyond average success rate or expected return, practical deployment also benefits from sufficient behavioral coverage: an agent should be able to produce a set of complementary high-quality strategies rather than relying on a single dominant path~\citep{batra2023proximal,cully2017quality}.
Such coverage is useful because the preferred strategy may depend on deployment constraints.
A conservative retrieve-then-verify workflow may be preferable when reliability is critical, while a more direct execution-oriented strategy may be preferable when latency or interaction cost is limited.
Moreover, when the environment changes (e.g., when an object is unavailable, a tool fails, or the accessible action set is restricted), an agent with multiple behaviorally distinct strategies is more likely to retain a viable alternative path.
This motivates diversity not merely as variation in final responses, but as task-relevant variation in the trajectories by which agents achieve their goals.

\begin{figure}[t]
    \centering
    \includegraphics[width=\linewidth]{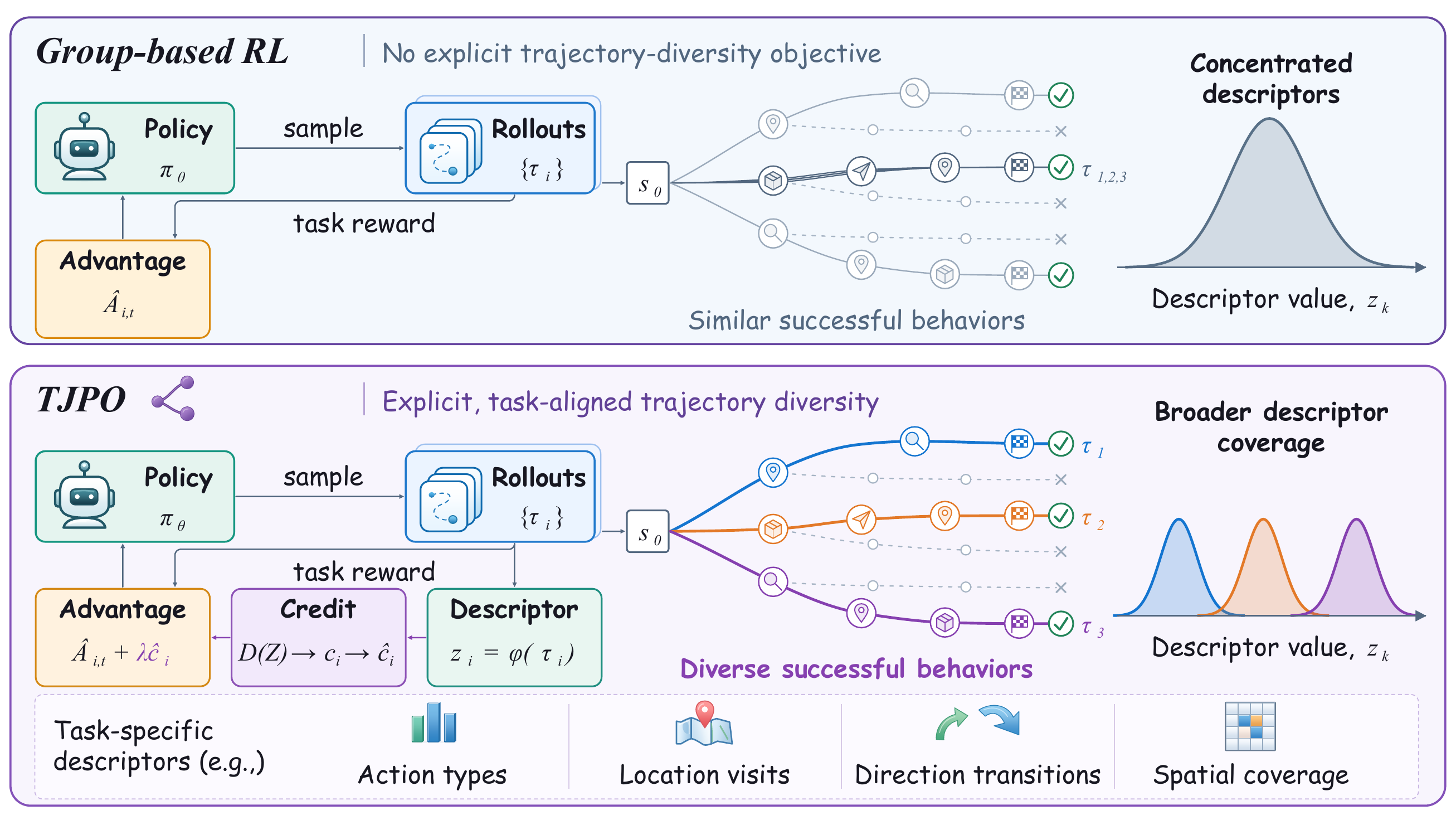}
    \caption{
        Motivation and schematic overview of TJPO.
        Without an explicit diversity objective, successful rollouts
        may concentrate on similar behaviors (top).
        TJPO augments the group-based advantage with descriptor-based
        trajectory-level diversity credits to encourage task-relevant
        behavioral variation (bottom).
    }
    \label{fig:motivation}
\end{figure}

However, such behavioral coverage does not follow automatically from RL-based post-training, where optimization primarily targets task rewards or preference alignment~\citep{bai2024digirl,feng2025towards,chen2025reinforcement,wang2025ragen,feng2505group}.
Without incentives to preserve diversity, policy optimization can favor a small set of reliably rewarded behaviors, while other valid strategies become less likely to be sampled and reinforced~\citep{orbert2024understand}.
Existing mechanisms can encourage diversity through decoding heuristics, entropy regularization, or other distribution-level regularizers~\citep{yao2025diversity,li2025choice}.
While these mechanisms can support exploration and response variation, encouraging variation alone does not specify \emph{what} should be diverse for a particular task.
These mechanisms typically operate on generic properties of the policy distribution rather than task-specific behavioral dimensions, leaving their effect on strategy-level diversity implicit.
Consequently, greater response variation does not necessarily translate into distinct reasoning structures, planning patterns, or tool-use behaviors that matter in practice.
The resulting diversity may therefore remain weakly aligned with task requirements and difficult to interpret or control.
This motivates a key question: how can we explicitly define and optimize task-relevant trajectory diversity in RL-based post-training while maintaining strong task performance?

To address this question, we propose \emph{Trajectory-guided Joint Policy Optimization} (TJPO), a simple framework for optimizing user-specified, task-specific trajectory diversity in RL-based post-training.
Our key idea is to represent each sampled trajectory using a set of trajectory descriptors chosen to reflect the behavioral variations that matter for the target environment, and to define diversity as a set-level functional over the resulting descriptor matrix.
This formulation makes the diversity objective explicit and controllable, rather than leaving diversity to emerge indirectly from generic stochasticity or response variation.
We further integrate this signal into group-based policy optimization by translating group-level diversity into trajectory-level credit, making TJPO readily compatible with GRPO-style and GiGPO-style training. Figure~\ref{fig:motivation} illustrates the motivation and
 diversity credit introduced by TJPO.

We evaluate TJPO on two complementary environments: Sokoban, a planning-intensive game environment, and ALFWorld, a long-horizon agentic environment~\citep{shridhar2020alfworld,schrader2018gymsokoban}.
For each environment, we instantiate user-specified, task-specific trajectory descriptors tailored to the corresponding notion of meaningful behavioral variation.
On both environments, TJPO improves explicit trajectory diversity over strong group-based RL baselines while maintaining competitive task performance.
These results suggest that task-specific trajectory descriptors provide a practical interface for optimizing meaningful behavioral diversity in RL-based post-training.
Our contributions are summarized as follows:

\begin{itemize}
    \item We formulate explicit trajectory diversity for RL-based post-training through user-specified, task-specific trajectory descriptors and set-level diversity functionals, providing an interpretable and controllable way to specify \emph{what} should be diverse.

    \item We introduce Trajectory-guided Joint Policy Optimization (TJPO), a single-policy framework that converts group-level diversity into trajectory-level learning signals, enabling explicit diversity optimization within group-based RL without maintaining a population of policies.

    \item We validate TJPO on ALFWorld and Sokoban, showing that it improves task-specific trajectory diversity while maintaining competitive task performance. Further trajectory-level and intervention-based evaluations show that the learned diversity corresponds to distinct successful strategies and can improve compliance with user constraints and robustness to environmental disruptions without additional training.
\end{itemize}


\section{Preliminary}

\subsection{RL for LLMs}
In RL-based post-training for LLMs, generation is modeled as a sequential decision-making process.
Given an input $x \in \mathcal{X}$, the model acts as a policy that autoregressively generates a trajectory $\tau=(\tau^1,\tau^2,\dots,\tau^T)\in\mathcal{T}$, where $\mathcal{T}$ denotes the space of all possible outputs.
The policy is parameterized by $\theta$ and factorizes as
\begin{equation}
    \pi_{\theta}(\tau \mid x)
    :=
    \prod_{t=1}^{T}
    \pi_{\theta}(\tau^t \mid x, \tau^{<t}),
\end{equation}
where $\tau^{<t}=(\tau^1,\dots,\tau^{t-1})$.
A reward function is then used to evaluate the generated trajectory and optimize the policy.

\subsection{Group-based Reinforcement Learning Algorithms}
\label{sec:grpo}

\paragraph{Group Relative Policy Optimization (GRPO)}
GRPO eliminates the need for a separate critic model by estimating the baseline from a group of sampled trajectories.
For each input $x$, it samples a group of trajectories $\{\tau_1,\dots,\tau_G\}$ from the old policy $\pi_{\mathrm{old}}$.
Writing the trajectory likelihood ratio as
$\rho_i(\theta)=\pi_{\theta}(\tau_i|x)/\pi_{\mathrm{old}}(\tau_i|x)$,
GRPO optimizes the current policy $\pi_\theta$ via
\begin{equation}
    \label{equ:grpo}
    \mathbb{E}
    \Bigg[
    \frac{1}{G}\sum_{i=1}^{G}
    \Bigg(
    \min \Big(
    \rho_i(\theta) A_i,
    \mathrm{clip}\big(\rho_i(\theta),1-\epsilon,1+\epsilon\big)A_i
    \Big)
    -\beta\mathbb{D}_{\mathrm{KL}}
    (\pi_{\theta}\|\pi_{\mathrm{ref}})
    \Bigg)
    \Bigg],
\end{equation}
where the expectation is over
$x\sim\mathcal{X}$ and
$\{\tau_i\}_{i=1}^G\sim\pi_{\mathrm{old}}(\cdot|x)$,
$\epsilon$ and $\beta$ are hyperparameters, and
\begin{equation}
    \mathbb{D}_{\mathrm{KL}}(\pi_{\theta}\|\pi_{\mathrm{ref}})
    =
    \frac{\pi_{\mathrm{ref}}(\tau_i|x)}{\pi_{\theta}(\tau_i|x)}
    -
    \log \frac{\pi_{\mathrm{ref}}(\tau_i|x)}{\pi_{\theta}(\tau_i|x)}
    -1,
\quad
    A_i
    =
    \frac{r_i - \mathrm{mean}(\{r_1,\dots,r_G\})}
    {\mathrm{std}(\{r_1,\dots,r_G\})}.
\end{equation}

\paragraph{Group-in-Group Policy Optimization (GiGPO)}
While GRPO is effective for single-turn generation, long-horizon agent tasks present a more challenging credit assignment problem due to sparse and delayed rewards.
GiGPO extends group-based RL by introducing an additional step-level relative advantage on top of the trajectory-level advantage used in GRPO.

To provide finer-grained supervision, GiGPO constructs step-level groups using \emph{anchor states}, i.e., repeated or aligned environment states observed across trajectories.
For an action taken at step $t$ in trajectory $i$, it computes a micro advantage by comparing it against other actions originating from the same anchor-state group:
\begin{equation}
    A_{i,t}^{\mathrm{micro}}
    =
    \frac{u_{i,t} - \mathrm{mean}(\mathcal{H}(s_{i,t}))}
    {\mathrm{std}(\mathcal{H}(s_{i,t}))},
\end{equation}
where $\mathcal{H}(s_{i,t})$ denotes the set of step-level utilities associated with the anchor-state group of $s_{i,t}$.
The final advantage combines the trajectory-level signal from GRPO with the step-level signal:
\begin{equation}
    \hat A_{i,t}
    =
    A_i + \omega A_{i,t}^{\mathrm{micro}},
\end{equation}
where $\omega$ controls the strength of the micro-level credit assignment.
GiGPO is particularly suitable for long-horizon LLM agent training, where useful supervision must be propagated not only across full trajectories but also across individual decision steps.

\section{Method}
\label{sec:method}

We study explicit trajectory diversity in group-based policy optimization for LLMs. Our goal is to encourage \emph{intent-aligned} behavioral variation, namely, variation along task-relevant dimensions specified by the practitioner, rather than generic randomness induced by token-level entropy or decoding noise.

\subsection{Problem Setup}
\label{subsec:setup}

Following the notation in Section~\ref{sec:grpo}, for each input $x$ we sample a group of trajectories
\begin{equation}
    \mathcal{G}(x) = \{\tau_i\}_{i=1}^G,
    \qquad
    \tau_i \sim \pi_\theta(\cdot \mid x).
\end{equation}
Each trajectory $\tau_i$ receives a task reward
    $r_i^{\mathrm{task}} = R(x,\tau_i)$.
Our goal is to augment this reward with an explicit diversity signal defined over the sampled trajectory group $\mathcal{G}(x)$.

\subsection{Intent-Aligned Trajectory Descriptors}
\label{subsec:descriptor}

\begin{definition}[Intent-Aligned Trajectory Descriptor]
Given the trajectory space $\mathcal{T}$, an intent-aligned trajectory descriptor is a user-specified scalar function
\begin{equation}
    \phi_k: \mathcal{T} \rightarrow \mathbb{R},
    \qquad k=1,\dots,m,
\end{equation}
that measures one task-relevant axis of behavioral variation.
\end{definition}

A descriptor family $\{\phi_k\}_{k=1}^m$ induces a trajectory-to-vector mapping
\begin{equation}
    \mathbf{z}(\tau)
    =
    \big[\phi_1(\tau), \dots, \phi_m(\tau)\big]^\top
    \in \mathbb{R}^m.
\end{equation}
For each sampled trajectory $\tau_i \in \mathcal{G}(x)$, we write
$
    \mathbf{z}_i = \mathbf{z}(\tau_i)
$. Collecting all descriptors in the group yields the descriptor matrix
\begin{equation}
    Z=[
    \mathbf{z}_1 
    \cdots 
    \mathbf{z}_G
    ]^\top
    \in \mathbb{R}^{G \times m}.
\end{equation}

By construction, the descriptor family $\{\phi_k\}_{k=1}^m$ is intent-aligned: it explicitly specifies which forms of trajectory variation are desirable for the task, rather than relying on task-agnostic statistics over the policy's state-action distribution.

\subsection{Group-Level Explicit Diversity}
\label{subsec:diversity}

We now define explicit diversity as a set-level function over the descriptor vectors induced by a sampled trajectory group.

\begin{definition}[Explicit Diversity Functional]
Given a descriptor matrix $Z \in \mathbb{R}^{G \times m}$, where each row corresponds to the descriptor vector of one sampled trajectory, an explicit diversity functional is a mapping
\begin{equation}
    \mathcal{D}: \mathbb{R}^{G \times m} \rightarrow \mathbb{R},
\end{equation}
which assigns a scalar diversity score to the trajectory group.
We require $\mathcal{D}$ to be permutation-invariant with respect to the ordering of trajectories, i.e., for any $G \times G$ permutation matrix $P$,
\begin{equation}
    \mathcal{D}(Z) = \mathcal{D}(PZ).
\end{equation}
\end{definition}

\paragraph{Pairwise diversity.}
A simple choice is the average pairwise dispersion:
\begin{equation}
    \mathcal{D}_{\mathrm{pair}}(Z)
    =
    \frac{2}{G(G-1)}
    \sum_{1 \le i < j \le G}
    d(\mathbf{z}_i,\mathbf{z}_j),
    \label{eq:pairwise_diversity}
\end{equation}
where $d(\cdot,\cdot)$ is a distance function, e.g., Euclidean distance or cosine distance.
By construction, $\mathcal{D}_{\mathrm{pair}}$ is permutation-invariant.

\paragraph{Determinantal diversity.}
To capture set-level coverage beyond pairwise comparisons, we also consider a determinantal form.
Let
\begin{equation}
    K_{ij} = k(\mathbf{z}_i,\mathbf{z}_j),
\end{equation}
where $k(\cdot,\cdot)$ is a positive semidefinite kernel, such as an RBF kernel.
We define
\begin{equation}
    \mathcal{D}_{\mathrm{det}}(Z)
    =
    \log \det (K + \eta I),
    \label{eq:det_diversity}
\end{equation}
where $\eta > 0$ ensures numerical stability.
Maximizing~\eqref{eq:det_diversity} encourages the sampled trajectories to span a larger volume in the descriptor space, thereby reducing redundancy.
This functional is also permutation-invariant, since permuting the rows of $Z$ only induces a simultaneous row-column permutation of $K$, which leaves the determinant unchanged.

The formulation above is intentionally modular: any permutation-invariant functional over the descriptor set can be plugged into our framework.

\subsection{From Group Diversity to Trajectory-Level Credit}
\label{subsec:credit}

A subtle but important issue arises when combining explicit diversity with GRPO-style training.
Since GRPO computes \emph{group-relative} advantages, adding the same group-level scalar bonus to every trajectory has no effect after baseline subtraction.
Indeed, the shared term in
\begin{equation}
\bar r_i = r_i^{\mathrm{task}} + \lambda \mathcal{D}(Z)
\qquad \forall i,
\end{equation}
vanishes after group centering, so diversity must be converted into
\emph{trajectory-level} credit.

For pairwise diversity, we use the average distance to the remaining group
members:
\begin{equation}
c_i^{\mathrm{pair}}(Z)
=
\frac{1}{G-1}
\sum_{j \ne i}
d(\mathbf{z}_i,\mathbf{z}_j).
\label{eq:pairwise_credit}
\end{equation}
As derived in Appendix~\ref{app:credit_derivation}, this credit is a positive affine
transformation of the exact leave-one-out change within a fixed group and
therefore yields identical group-standardized values.

For determinantal diversity, with $A=K+\eta I$, we use
\begin{equation}
c_i^{\mathrm{det}}(Z)
=
\frac{\det(A)}
{\det(A_{-i,-i})}
=
\exp\!\left(
\mathcal{D}_{\mathrm{det}}(Z)-\mathcal{D}_{\mathrm{det}}(Z_{-i})
\right).
\label{eq:det_credit}
\end{equation}
This is the exponential of the leave-one-out log-determinant change.

Finally, we standardize the credits within each sampled group and add them
to the GiGPO advantage:
\begin{align}
\hat c_i
&=
\frac{c_i-\bar c}
{\max\{\operatorname{std}(c),\epsilon\}},
\label{eq:standardized_diversity_credit}\\
\widetilde A_{i,t}
&=
\hat A_{i,t}
+
\lambda \hat c_i,
\label{eq:augmented_advantage}
\end{align}
where $\epsilon>0$ handles near-zero variance and $\lambda\geq0$ controls
the task--diversity trade-off. The trajectory-level credit is shared across
the valid response tokens associated with trajectory $i$.

Our method preserves the simplicity of GRPO-style training while explicitly optimizing task-relevant trajectory diversity.
The key difference from implicit diversity regularizers is that diversity is defined over full-trajectory descriptors chosen to match the intended behavioral variation of each environment, rather than generic statistics of token-level stochasticity.

Taken together, our framework cleanly separates \emph{what} should be diverse from \emph{how} diversity is optimized.
The descriptor family $\{\phi_k\}$ specifies the intended axes of variation, while the diversity functional $\mathcal{D}$ specifies how a group of sampled trajectories should spread out in the resulting descriptor space.
This modular design allows the same optimization framework to be instantiated across diverse environments with task-specific notions of meaningful diversity.
We term our method \emph{Trajectory-guided Joint Policy Optimization} (TJPO).

\section{Experiments}
\label{sec:exp}
We evaluate TJPO through the following questions:

\begin{enumerate}[Q1.]
    \item Can TJPO improve task-specific diversity while maintaining strong task performance?
    \item How do trajectory descriptors and diversity functionals shape the learned behavioral variation?
    \item How does the diversity coefficient $\lambda$ affect the diversity--performance trade-off?
    \item Does the resulting behavioral diversity provide practical benefits under user constraints and environmental disruptions?
\end{enumerate}

\subsection{Experimental Setup}
\label{subsec:exp_setup}

\paragraph{Environments.}
We evaluate on ALFWorld~\citep{shridhar2020alfworld} and Sokoban~\citep{schrader2018gymsokoban}.
ALFWorld requires an agent to complete household manipulation tasks through multi-step text interactions, where trajectories consist of observations, reasoning, and textual actions.
Sokoban is a grid-based box-pushing puzzle in which the agent must plan action sequences to push boxes onto target cells.
We provide detailed environment descriptions in Appendix~\ref{app:env_details}.

\paragraph{Models.}
We use Qwen2.5-7B-Instruct as the base model for ALFWorld and Qwen3-VL-4B-Instruct for Sokoban.
Within each environment, all trained policies start from the same base model and use the same rollout budget, optimization hyperparameters, and evaluation protocol.
Our decoding-only baseline reuses the trained GiGPO checkpoint without any additional optimization.

\paragraph{Training backbone.}
We implement our method on top of GiGPO~\citep{feng2505group}, a group-based RL algorithm for long-horizon LLM agent training.
For each input instance, the policy samples a group of trajectories.
We compute a trajectory-level diversity credit for each trajectory,
standardize the credits within the sampled group, and add the resulting
signal to the GiGPO advantage as described in
Equation~\ref{eq:augmented_advantage}.
Implementation details are provided in Appendix~\ref{app:training_details}.

\paragraph{Evaluation metrics.}
We report both task performance and explicit diversity.
For task performance, we use success rate and average trajectory reward.
For diversity, we sample $N_{\mathrm{eval}}$ trajectories per input and compute descriptor-based diversity over the successful subset of sampled trajectories using the corresponding evaluation descriptor.

\paragraph{Baselines.}
Our primary baseline is GiGPO without explicit diversity optimization.
For ALFWorld, we also evaluate GiGPO+PP and GiGPO+RP, inference-only baselines
that apply a presence penalty of $0.5$ or repetition penalty of $1.05$ to the
same checkpoint while leaving all other evaluation settings unchanged.
We compare action-type and location descriptors with pairwise or determinantal diversity.
For Sokoban, we compare GiGPO with two task-specific descriptors, denoted as direction-transition and spatial-coverage, which are detailed in Appendix~\ref{app:bd_details}.

\subsection{Main Results on ALFWorld}
\label{subsec:alfworld_results}

\begin{table}[t]
\centering
\begin{minipage}[t]{0.48\textwidth}
\centering
\setlength{\tabcolsep}{3pt}
\caption{Pairwise diversity on ALFWorld.}
\label{tab:alfworld_main}
\begin{tabular}{@{}lccc@{}}
\toprule
\textbf{Method}
& \textbf{Success} $\uparrow$
& $\boldsymbol{D_{\mathrm{act}}^{\mathrm{pair}}}$ $\uparrow$
& $\boldsymbol{D_{\mathrm{loc}}^{\mathrm{pair}}}$ $\uparrow$ \\
\midrule
GiGPO
& \texttt{0.9160}
& \texttt{0.0495}
& \texttt{0.0858} \\
GiGPO + PP
& \texttt{0.9082}
& \texttt{0.0458}
& \texttt{0.0895} \\
GiGPO + RP
& \texttt{0.9229}
& \texttt{0.0494}
& \texttt{0.0873} \\
\midrule
TJPO$_{\mathrm{act}}^{\mathrm{pair}}$
& \texttt{0.8682}
& \textbf{\texttt{0.0780}}
& \texttt{0.1002} \\
TJPO$_{\mathrm{loc}}^{\mathrm{pair}}$
& \texttt{0.9053}
& \texttt{0.0601}
& \textbf{\texttt{0.1316}} \\
\bottomrule
\end{tabular}
\end{minipage}\hfill%
\begin{minipage}[t]{0.48\textwidth}
\centering
\setlength{\tabcolsep}{3pt}
\caption{DPP diversity ($\times 1000$) on ALFWorld.}
\label{tab:alfworld_main_dpp}
\begin{tabular}{@{}lccc@{}}
\toprule
\textbf{Method}
& \textbf{Success} $\uparrow$
& \shortstack{
    $\boldsymbol{D_{\mathrm{act}}^{\mathrm{dpp}}}$ $\uparrow$
  }
& \shortstack{
    $\boldsymbol{D_{\mathrm{loc}}^{\mathrm{dpp}}}$ $\uparrow$
  } \\
\midrule
GiGPO
& \texttt{0.9160}
& \texttt{1.887}
& \texttt{2.500} \\
GiGPO + PP
& \texttt{0.9082}
& \texttt{1.858}
& \texttt{2.063} \\
GiGPO + RP
& \texttt{0.9229}
& \texttt{1.135}
& \texttt{1.544} \\
\midrule
TJPO$_{\mathrm{act}}^{\mathrm{dpp}}$
& \texttt{0.9160}
& \textbf{\texttt{4.100}}
& \texttt{6.655} \\
TJPO$_{\mathrm{loc}}^{\mathrm{dpp}}$
& \texttt{0.8779}
& \texttt{3.058}
& \textbf{\texttt{7.940}} \\
\bottomrule
\end{tabular}
\end{minipage}
\end{table}


Tables~\ref{tab:alfworld_main} and~\ref{tab:alfworld_main_dpp}
report the pairwise and determinantal diversity results on ALFWorld, respectively.
All TJPO variants use $\lambda=0.3$.
For determinantal diversity $D^{\mathrm{dpp}}$, we use the determinant-root
score defined in Appendix~\ref{app:alfworld_determinantal_eval},
with values reported in units of $10^{-3}$.
We compare standard GiGPO, two decoding-only penalty baselines, and four TJPO
variants obtained by combining two task-specific trajectory descriptors,
action type and location, with either pairwise or determinantal diversity. The results show that TJPO improves explicit trajectory diversity under the intended descriptor metrics while maintaining competitive task performance.
For the action-type descriptor, TJPO with pairwise diversity increases
$D_{\mathrm{act}}^{\mathrm{pair}}$ from $0.0495$ to $0.0780$,
with a success rate of $0.8682$; the determinantal variant increases
$D_{\mathrm{act}}^{\mathrm{dpp}}$ from $1.887\times10^{-3}$ to
$4.100\times10^{-3}$ with a success rate of $0.9160$.
For the location descriptor, TJPO with pairwise diversity increases
$D_{\mathrm{loc}}^{\mathrm{pair}}$ from $0.0858$ to $0.1316$,
with a success rate of $0.9053$; the determinantal variant increases
$D_{\mathrm{loc}}^{\mathrm{dpp}}$ from $2.500\times10^{-3}$ to
$7.940\times10^{-3}$, with a success rate of $0.8779$. These results indicate that user-specified descriptors provide a controllable interface for shaping which aspects of trajectory behavior become diverse.
The cross-metric results show that the descriptor determines
which aspects of behavior become more diverse.
Action-type optimization primarily changes the composition of action categories, whereas location-level optimization encourages broader variation in location visitation patterns. 

\begin{figure}[t]
\centering
\subfloat[GiGPO navigation strategies.\label{fig:location_routes_gigpo}]{%
    \includegraphics[width=0.48\textwidth]{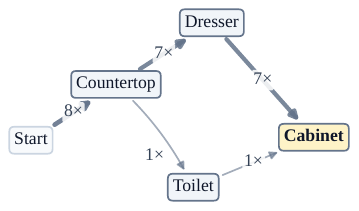}}
\hfill
\subfloat[TJPO navigation strategies.\label{fig:location_routes_tjpo}]{%
    \includegraphics[width=0.48\textwidth]{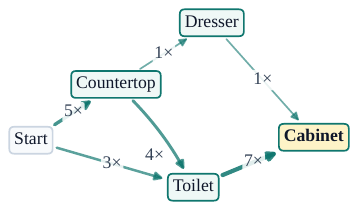}}
\caption{
Successful navigation strategies for ``put a spraybottle in cabinet''.
TJPO uses the location descriptor with pairwise diversity and $\lambda=0.3$.
Both methods succeed in all eight rollouts.
Nodes denote visited locations; edge labels and widths encode navigation transition counts across the eight rollouts.
}
\label{fig:alfworld_location_evidence}
\end{figure}


\subsection{Main Results on Sokoban}
\label{subsec:sokoban_results}

\begin{table}[t]
\centering
\begin{minipage}[t]{0.48\textwidth}
\vspace{0pt}
\centering
\setlength{\tabcolsep}{3pt}
\caption{
Main results on Sokoban.
}
\label{tab:sokoban_main}
\begin{tabular}{@{}lccc@{}}
\toprule
\textbf{Method}
& \textbf{Success} $\uparrow$
& $\boldsymbol{D_{\mathrm{dir}}^{\mathrm{pair}}}$ $\uparrow$
& $\boldsymbol{D_{\mathrm{spa}}^{\mathrm{pair}}}$ $\uparrow$ \\
\midrule
GiGPO
& \texttt{0.8281}
& \texttt{0.0146}
& \texttt{0.0092} \\
TJPO$_{\mathrm{dir}}^{\mathrm{pair}}$
& \texttt{0.8428}
& \textbf{\texttt{0.0675}}
& \texttt{0.0503} \\
TJPO$_{\mathrm{spa}}^{\mathrm{pair}}$
& \texttt{0.8057}
& \texttt{0.0595}
& \textbf{\texttt{0.0518}} \\
\bottomrule
\end{tabular}
\end{minipage}\hfill%
\begin{minipage}[t]{0.48\textwidth}
\vspace{0pt}
\centering
\setlength{\tabcolsep}{2pt}
\caption{
Ablation on the $\lambda$ on ALFWorld.
}
\label{tab:lambda_ablation_location}
\begin{tabular}{@{}lcccc@{}}
\toprule
\textbf{Method}
& $\boldsymbol{\lambda}$
& \textbf{Success} $\uparrow$
& $D_{\mathrm{loc}}^{\mathrm{pair}}$ $\uparrow$
& \shortstack{\textbf{Avg. Steps} $\downarrow$} \\
\midrule
TJPO & 0
& \texttt{0.9160}
& \texttt{0.0858}
& \texttt{9.2217} \\
TJPO & 0.1
& \texttt{0.9209}
& \texttt{0.0942}
& \texttt{9.5811} \\
TJPO & 0.3
& \texttt{0.9053}
& \texttt{0.1316}
& \texttt{9.6947} \\
TJPO & 0.5
& \texttt{0.9004}
& \texttt{0.1356}
& \texttt{8.1150} \\
\bottomrule
\end{tabular}
\end{minipage}
\end{table}


Table~\ref{tab:sokoban_main} reports the main results on Sokoban. All TJPO variants use $\lambda=0.3$ and pairwise diversity.
Compared with the GiGPO baseline, TJPO improves explicit trajectory diversity under both task-specific descriptors.
When optimizing the direction-transition descriptor, TJPO increases $D_{\mathrm{dir}}^{\mathrm{pair}}$ from $0.0146$ to $0.0675$, with a success rate of $0.8428$.
When optimizing the spatial-coverage descriptor, TJPO increases $D_{\mathrm{spa}}^{\mathrm{pair}}$ from $0.0092$ to $0.0518$, with a success rate of $0.8057$.
The results show that TJPO can substantially improve planning-level trajectory diversity while maintaining competitive task performance. Together with the ALFWorld results in Section~\ref{subsec:alfworld_results},
these findings suggest that our framework is not limited to tool-use
trajectories, but also applies to planning-intensive game environments
where meaningful diversity is reflected in movement patterns and
state-space coverage.

\subsection{Case Study}
\label{subsec:alfworld_case_study}
Figure~\ref{fig:alfworld_location_evidence} compares the successful navigation strategies of GiGPO and TJPO for the task ``put a spraybottle in cabinet''. GiGPO follows the same route in seven of eight rollouts.
TJPO instead distributes its rollouts across three successful navigation strategies, including a more direct route through the Toilet. 
This example illustrates that explicit location descriptor encourages
alternative efficient solution modes rather than merely increasing a scalar
diversity score. We further expand the route-level case study in
Appendix~\ref{app:location_case_study}.

\subsection{Ablation on Diversity Weight}
\label{subsec:lambda_ablation}

Table~\ref{tab:lambda_ablation_location} studies the effect of the diversity coefficient $\lambda$.
For the location descriptor, increasing $\lambda$ improves $D_{\mathrm{loc}}^{\mathrm{pair}}$, from $0.0858$ at $\lambda=0$ to $0.1356$ at $\lambda=0.5$, while maintaining competitive task performance.
The mean successful-trajectory length remains comparable at $\lambda \leq 0.3$ and decreases to $8.1150$ steps at $\lambda=0.5$, indicating that the diversity gain does not result from longer successful interactions.

\subsection{Practical Benefits of Behavioral Diversity}
\label{subsec:location_access}

Beyond diversity scores, we examine whether behavioral diversity helps
agents satisfy user-specific requirements and cope with environmental
disruptions. We use location access in ALFWorld as a concrete example:
\emph{User Constraints} specifies locations the agent must avoid,
whereas \emph{Disruption} makes task-relevant locations
unavailable. \emph{Standard} denotes the original task without either
intervention. We compare GiGPO with location-pairwise TJPO at $\lambda=0.5$
on tasks with verified alternative solutions.
Details are provided in Appendix~\ref{app:known_location_protocol}.

\begin{table}[t]
\centering
\caption{
Practical benefits of behavioral diversity on ALFWorld.
}
\label{tab:location_access}
\begin{tabular}{lccc}
\toprule
\textbf{Method}
& \textbf{Standard Success} $\uparrow$
& \textbf{User Constraints Success} $\uparrow$
& \textbf{Disruption Success} $\uparrow$ \\
\midrule
GiGPO
& \textbf{\texttt{0.9917}}
& \texttt{0.7708}
& \texttt{0.7792} \\
TJPO
& \texttt{0.9750}
& \textbf{\texttt{0.8542}}
& \textbf{\texttt{0.8542}} \\
\bottomrule
\end{tabular}
\end{table}

Table~\ref{tab:location_access} shows that TJPO improves compliant success
under user constraints from $77.08\%$ to $85.42\%$ and task success under
environmental disruption from $77.92\%$ to $85.42\%$.
These results illustrate the practical value of behavioral diversity in
this setting: producing successful behaviors that meet user requirements
and maintaining task completion when the environment is disrupted.

\section{Related work}

\paragraph{Diversity in RL}
Prior work on policy diversity in deep reinforcement learning can be broadly organized according to the role that diversity plays in the learning process~\citep{wu2023quality}.
One line of work leverages diversity to facilitate exploration, where diverse behaviors are encouraged mainly as an intermediate mechanism for improving final task performance~\citep{hong2018diversity, eysenbach2018diversity, parker2020effective, conti2018improving, peng2020non, yao2023policy}.
A second line incorporates diversity through constrained optimization, either by maximizing task quality under diversity constraints or by maximizing diversity while enforcing quality constraints~\citep{masood2019diversity, zhang2019learning, ghasemi2021multiple, zahavy2021discovering, zhou2022continuously,zhang2024two,zhang2026semantics}.
A third line jointly optimizes quality and diversity, commonly referred to as Quality-Diversity RL~\citep{cideron2020qd, pierrot2022diversity, tjanaka2022approximating, batra2023proximal,yang2025diverse}.
Our work extends the second paradigm to RL for LLM training.

\paragraph{Diversity in LLMs}
Prior work has explored diversity in LLMs from several perspectives, including supervised fine-tuning, preference optimization, RL-based post-training, diversity evaluation. 
In supervised and preference-based post-training, existing methods aim to preserve or improve response diversity during alignment, often by modifying the training objective, constructing diverse preference pairs, or encouraging diverse high-quality generations~\citep{li2025preserving, lanchantin2025diverse, chung2025modifying, chen2025post,wang2026learning}. 
Complementary studies evaluate linguistic, semantic, and epistemic diversity in LLM outputs, showing that diversity is an important dimension beyond average quality or accuracy~\citep{shypula2025evaluating, guo2025benchmarking, wright2025epistemic}. 
In RL-related settings, recent work has begun to study diversity in LLM reasoning, while other approaches use divergence choices or entropy-based objectives to indirectly preserve exploration and mitigate diversity collapse~\citep{yao2025diversity, li2025choice}. 
These works are closely related to ours, but they typically define or encourage diversity through variation among token-level diversity, embedding-based semantic diversity, or implicit distributional regularization.
In contrast, our work focuses on \emph{explicit trajectory diversity}: diversity is defined through user-specified, task-specific trajectory descriptors that directly encode the intended behavioral variations of each environment, such as reasoning structures, planning patterns, or tool-use trajectories, and is optimized within RL-based post-training.

\section{Conclusion, Limitations, and Discussion}
\label{sec:conclusion}

We study explicit trajectory diversity in RL-based post-training for LLMs.
Unlike implicit notions of diversity that arise from sampling randomness, entropy regularization, or distributional constraints, our framework defines diversity through user-specified, task-specific trajectory descriptors.
These descriptors make explicit which behavioral dimensions should vary for a given environment, such as planning patterns and tool-use trajectories.
Building on this formulation, we introduce a framework to optimize diversity over sampled trajectory groups. By converting set-level diversity into trajectory-level learning signals, the proposed method can be integrated into group-based RL training, such as GiGPO.
Across Sokoban and agentic tasks, our experiments show that explicit trajectory diversity can be consistently improved while maintaining strong task performance.

While our results are encouraging, several aspects merit further study.
First, our framework relies on task-specific trajectory descriptors.
This design is intentional, since explicit diversity should reflect the behavioral variations that matter for the target environment.
At the same time, different descriptor choices may emphasize different aspects of behavior, and developing better interfaces or reusable descriptor templates across task families, such as reasoning, planning, tool use, and embodied interaction, would make the framework easier to apply while preserving its user-controlled nature.
Second, although our experiments span planning and agentic tasks, broader evaluation on richer agent settings would further clarify the generality of explicit trajectory diversity.
Promising directions include larger-scale tool ecosystems, open-ended web environments, multi-agent settings, longer horizons, larger action spaces, and more complex environmental shifts.
Overall, our results suggest that explicitly defining and optimizing task-aligned trajectory diversity is a useful step toward more controllable, robust, and adaptable LLM agents.

\bibliographystyle{plainnat}
\bibliography{references}

@article{guo2025deepseek,
  title={Deepseek-r1: Incentivizing reasoning capability in llms via reinforcement learning},
  author={Guo, Daya and Yang, Dejian and Zhang, Haowei and Song, Junxiao and Zhang, Ruoyu and Xu, Runxin and Zhu, Qihao and Ma, Shirong and Wang, Peiyi and Bi, Xiao and others},
  journal={arXiv preprint arXiv:2501.12948},
  year={2025}
}

@inproceedings{li2025preserving,
  title={Preserving diversity in supervised fine-tuning of large language models},
  author={Li, Ziniu and Chen, Congliang and Xu, Tian and Qin, Zeyu and Xiao, Jiancong and Luo, Zhi-Quan and Sun, Ruoyu},
  booktitle={The Thirteenth International Conference on Learning Representations},
  year={2025}
}

@article{li2025limr,
  title={Limr: Less is more for rl scaling},
  author={Li, Xuefeng and Zou, Haoyang and Liu, Pengfei},
  journal={arXiv preprint arXiv:2502.11886},
  year={2025}
}

@article{he2025deepmath,
  title={DeepMath-103K: A Large-Scale, Challenging, Decontaminated, and Verifiable Mathematical Dataset for Advancing Reasoning},
  author={He, Zhiwei and Liang, Tian and Xu, Jiahao and Liu, Qiuzhi and Chen, Xingyu and Wang, Yue and Song, Linfeng and Yu, Dian and Liang, Zhenwen and Wang, Wenxuan and others},
  journal={arXiv preprint arXiv:2504.11456},
  year={2025}
}

@article{yu2503dapo,
  title={Dapo: An open-source llm reinforcement learning system at scale, 2025},
  author={Yu, Qiying and Zhang, Zheng and Zhu, Ruofei and Yuan, Yufeng and Zuo, Xiaochen and Yue, Yu and Fan, Tiantian and Liu, Gaohong and Liu, Lingjun and Liu, Xin and others},
  journal={URL https://arxiv. org/abs/2503.14476},
  year={2025}
}

@misc{hu2025openreasonerzeroopensourceapproach,
      title={Open-Reasoner-Zero: An Open Source Approach to Scaling Up Reinforcement Learning on the Base Model}, 
      author={Jingcheng Hu and Yinmin Zhang and Qi Han and Daxin Jiang and Xiangyu Zhang and Heung-Yeung Shum},
      year={2025},
      eprint={2503.24290},
      archivePrefix={arXiv},
      primaryClass={cs.LG},
      url={https://arxiv.org/abs/2503.24290}, 
}

@misc{curatedthoughts,
  author = {Hochlehnert, Andreas and Bhatnagar, Hardik and Udandarao, Vishaal and Prabhu, Ameya and Bethge, Matthias},
  title  = {CuratedThoughts: Data Curation for RL Training Datasets},
  url    = {https://huggingface.co/datasets/bethgelab/CuratedThoughts},
  year   = {2025},
}

@misc{zeng2025simplerl,
  title={7B Model and 8K Examples: Emerging Reasoning with Reinforcement Learning is Both Effective and Efficient},
  author={Weihao Zeng and Yuzhen Huang and Wei Liu and Keqing He and Qian Liu and Zejun Ma and Junxian He},
  year={2025},
  howpublished={\url{https://hkust-nlp.notion.site/simplerl-reason}},
  note={Notion Blog}
}

@article{kazemnejad2024vineppo,
  title={Vineppo: Unlocking rl potential for llm reasoning through refined credit assignment},
  author={Kazemnejad, Amirhossein and Aghajohari, Milad and Portelance, Eva and Sordoni, Alessandro and Reddy, Siva and Courville, Aaron and Roux, Nicolas Le},
  journal={arXiv preprint arXiv:2410.01679},
  year={2024}
}

@article{yuan2025vapo,
  title={VAPO: Efficient and reliable reinforcement learning for advanced reasoning tasks},
  author={Yuan, Yufeng and Yu, Qiying and Zuo, Xiaochen and Zhu, Ruofei and Xu, Wenyuan and Chen, Jiaze and Wang, Chengyi and Fan, TianTian and Du, Zhengyin and Wei, Xiangpeng and others},
  journal={arXiv preprint arXiv:2504.05118},
  year={2025}
}

@article{yuan2025s,
  title={What's Behind PPO's Collapse in Long-CoT? Value Optimization Holds the Secret},
  author={Yuan, Yufeng and Yue, Yu and Zhu, Ruofei and Fan, Tiantian and Yan, Lin},
  journal={arXiv preprint arXiv:2503.01491},
  year={2025}
}

@article{liu2025understanding,
  title={Understanding r1-zero-like training: A critical perspective},
  author={Liu, Zichen and Chen, Changyu and Li, Wenjun and Qi, Penghui and Pang, Tianyu and Du, Chao and Lee, Wee Sun and Lin, Min},
  journal={arXiv preprint arXiv:2503.20783},
  year={2025}
}

@article{zhang2025srpo,
  title={Srpo: A cross-domain implementation of large-scale reinforcement learning on llm},
  author={Zhang, Xiaojiang and Wang, Jinghui and Cheng, Zifei and Zhuang, Wenhao and Lin, Zheng and Zhang, Minglei and Wang, Shaojie and Cui, Yinghan and Wang, Chao and Peng, Junyi and others},
  journal={arXiv preprint arXiv:2504.14286},
  year={2025}
}

@inproceedings{yang2025diverse,
  title={Diverse policies recovering via pointwise mutual information weighted imitation learning},
  author={Yang, Hanlin and Yao, Jian and Liu, Weiming and Wang, Qing and Qin, Hanmin and Tang, Kirk and Xiong, Jiechao and Yu, Chao and Li, Kai and Xing, Junliang and others},
  booktitle={International Conference on Learning Representations},
  volume={2025},
  pages={24860--24878},
  year={2025}
}

@article{hu2025reinforce++,
  title={Reinforce++: A simple and efficient approach for aligning large language models},
  author={Hu, Jian},
  journal={arXiv preprint arXiv:2501.03262},
  year={2025}
}

@inproceedings{wu2023quality,
  title={Quality-similar diversity via population based reinforcement learning},
  author={Wu, Shuang and Yao, Jian and Fu, Haobo and Tian, Ye and Qian, Chao and Yang, Yaodong and Fu, Qiang and Wei, Yang},
  booktitle={The eleventh international conference on learning representations},
  year={2023}
}

@article{hong2018diversity,
  title={Diversity-driven exploration strategy for deep reinforcement learning},
  author={Hong, Zhang-Wei and Shann, Tzu-Yun and Su, Shih-Yang and Chang, Yi-Hsiang and Fu, Tsu-Jui and Lee, Chun-Yi},
  journal={Advances in neural information processing systems},
  volume={31},
  year={2018}
}

@article{eysenbach2018diversity,
  title={Diversity is all you need: Learning skills without a reward function},
  author={Eysenbach, Benjamin and Gupta, Abhishek and Ibarz, Julian and Levine, Sergey},
  journal={arXiv preprint arXiv:1802.06070},
  year={2018}
}

@article{parker2020effective,
  title={Effective diversity in population based reinforcement learning},
  author={Parker-Holder, Jack and Pacchiano, Aldo and Choromanski, Krzysztof M and Roberts, Stephen J},
  journal={Advances in Neural Information Processing Systems},
  volume={33},
  pages={18050--18062},
  year={2020}
}

@article{conti2018improving,
  title={Improving exploration in evolution strategies for deep reinforcement learning via a population of novelty-seeking agents},
  author={Conti, Edoardo and Madhavan, Vashisht and Petroski Such, Felipe and Lehman, Joel and Stanley, Kenneth and Clune, Jeff},
  journal={Advances in neural information processing systems},
  volume={31},
  year={2018}
}

@article{peng2020non,
  title={Non-local policy optimization via diversity-regularized collaborative exploration},
  author={Peng, Zhenghao and Sun, Hao and Zhou, Bolei},
  journal={arXiv preprint arXiv:2006.07781},
  year={2020}
}

@article{masood2019diversity,
  title={Diversity-inducing policy gradient: Using maximum mean discrepancy to find a set of diverse policies},
  author={Masood, Muhammad A and Doshi-Velez, Finale},
  journal={arXiv preprint arXiv:1906.00088},
  year={2019}
}

@inproceedings{zhang2019learning,
  title={Learning novel policies for tasks},
  author={Zhang, Yunbo and Yu, Wenhao and Turk, Greg},
  booktitle={International Conference on Machine Learning},
  pages={7483--7492},
  year={2019},
  organization={PMLR}
}

@inproceedings{ghasemi2021multiple,
  title={Multiple Plans are Better than One: Diverse Stochastic Planning},
  author={Ghasemi, Mahsa and Crafts, Evan Scope and Zhao, Bo and Topcu, Ufuk},
  booktitle={Proceedings of the International Conference on Automated Planning and Scheduling},
  volume={31},
  pages={140--148},
  year={2021}
}

@article{zahavy2021discovering,
  title={Discovering diverse nearly optimal policies with successor features},
  author={Zahavy, Tom and O'Donoghue, Brendan and Barreto, Andre and Mnih, Volodymyr and Flennerhag, Sebastian and Singh, Satinder},
  journal={arXiv preprint arXiv:2106.00669},
  year={2021}
}

@article{zhou2022continuously,
  title={Continuously discovering novel strategies via reward-switching policy optimization},
  author={Zhou, Zihan and Fu, Wei and Zhang, Bingliang and Wu, Yi},
  journal={arXiv preprint arXiv:2204.02246},
  year={2022}
}

@article{cideron2020qd,
  title={QD-RL: Efficient Mixing of Quality and Diversity in Reinforcement Learning. CoRR abs/2006.08505 (2020)},
  author={Cideron, Geoffrey and Pierrot, Thomas and Perrin, Nicolas and Beguir, Karim and Sigaud, Olivier},
  journal={arXiv preprint arXiv:2006.08505},
  year={2020}
}

@inproceedings{pierrot2022diversity,
  title={Diversity policy gradient for sample efficient quality-diversity optimization},
  author={Pierrot, Thomas and Mac{\'e}, Valentin and Chalumeau, Felix and Flajolet, Arthur and Cideron, Geoffrey and Beguir, Karim and Cully, Antoine and Sigaud, Olivier and Perrin-Gilbert, Nicolas},
  booktitle={Proceedings of the Genetic and Evolutionary Computation Conference},
  pages={1075--1083},
  year={2022}
}

@inproceedings{tjanaka2022approximating,
  title={Approximating gradients for differentiable quality diversity in reinforcement learning},
  author={Tjanaka, Bryon and Fontaine, Matthew C and Togelius, Julian and Nikolaidis, Stefanos},
  booktitle={Proceedings of the Genetic and Evolutionary Computation Conference},
  pages={1102--1111},
  year={2022}
}

@article{batra2023proximal,
  title={Proximal policy gradient arborescence for quality diversity reinforcement learning},
  author={Batra, Sumeet and Tjanaka, Bryon and Fontaine, Matthew C and Petrenko, Aleksei and Nikolaidis, Stefanos and Sukhatme, Gaurav},
  journal={arXiv preprint arXiv:2305.13795},
  year={2023}
}

@article{team2025kimi,
  title={Kimi k1.5: Scaling reinforcement learning with llms},
  author={Team, Kimi and Du, Angang and Gao, Bofei and Xing, Bowei and Jiang, Changjiu and Chen, Cheng and Li, Cheng and Xiao, Chenjun and Du, Chenzhuang and Liao, Chonghua and others},
  journal={arXiv preprint arXiv:2501.12599},
  year={2025}
}

@misc{openai-o1,
  title={Learning to reason with LLMs},
    author={OpenAI},
  howpublished={\url{https://openai.com/index/learning-to-reason-with-llms}},
  year={2024}
}

@inproceedings{meng2023deepscaler,
  title={Deepscaler: Holistic autoscaling for microservices based on spatiotemporal gnn with adaptive graph learning},
  author={Meng, Chunyang and Song, Shijie and Tong, Haogang and Pan, Maolin and Yu, Yang},
  booktitle={2023 38th IEEE/ACM International Conference on Automated Software Engineering (ASE)},
  pages={53--65},
  year={2023},
  organization={IEEE}
}

@article{Chu_Huang_Zhang_Wei_Wang_2025,
 title={GPG: A Simple and Strong Reinforcement Learning Baseline for Model Reasoning}, 
 author={Chu, Xiangxiang and Huang, Hailang and Zhang, Xiao and Wei, Fei and Wang, Yong}, 
journal={arXiv preprint arXiv:2504.02546},
 year={2025}, 
}

@article{yao2023policy,
  title={Policy space diversity for non-transitive games},
  author={Yao, Jian and Liu, Weiming and Fu, Haobo and Yang, Yaodong and McAleer, Stephen and Fu, Qiang and Yang, Wei},
  journal={Advances in Neural Information Processing Systems},
  volume={36},
  pages={67771--67793},
  year={2023}
}

@article{lanchantin2025diverse,
  title={Diverse preference optimization},
  author={Lanchantin, Jack and Chen, Angelica and Dhuliawala, Shehzaad and Yu, Ping and Weston, Jason and Sukhbaatar, Sainbayar and Kulikov, Ilia},
  journal={arXiv preprint arXiv:2501.18101},
  year={2025}
}

@article{chung2025modifying,
  title={Modifying large language model post-training for diverse creative writing},
  author={Chung, John Joon Young and Padmakumar, Vishakh and Roemmele, Melissa and Sun, Yuqian and Kreminski, Max},
  journal={arXiv preprint arXiv:2503.17126},
  year={2025}
}

@article{chen2025post,
  title={Post-training Large Language Models for Diverse High-Quality Responses},
  author={Chen, Yilei and Chakraborty, Souradip and Wolf, Lorenz and Paschalidis, Yannis and Pacchiano, Aldo},
  journal={arXiv preprint arXiv:2509.04784},
  year={2025}
}

@article{shypula2025evaluating,
  title={Evaluating the diversity and quality of llm generated content},
  author={Shypula, Alexander and Li, Shuo and Zhang, Botong and Padmakumar, Vishakh and Yin, Kayo and Bastani, Osbert},
  journal={arXiv preprint arXiv:2504.12522},
  year={2025}
}

@article{guo2025benchmarking,
  title={Benchmarking linguistic diversity of large language models},
  author={Guo, Yanzhu and Shang, Guokan and Clavel, Chlo{\'e}},
  journal={Transactions of the Association for Computational Linguistics},
  volume={13},
  pages={1507--1526},
  year={2025},
  publisher={MIT Press 255 Main Street, 9th Floor, Cambridge, Massachusetts 02142, USA~…}
}

@article{wright2025epistemic,
  title={Epistemic diversity and knowledge collapse in large language models},
  author={Wright, Dustin and Masud, Sarah and Moore, Jared and Yadav, Srishti and Antoniak, Maria and Christensen, Peter Ebert and Park, Chan Young and Augenstein, Isabelle},
  journal={arXiv preprint arXiv:2510.04226},
  year={2025}
}

@article{yao2025diversity,
  title={Diversity-aware policy optimization for large language model reasoning},
  author={Yao, Jian and Cheng, Ran and Wu, Xingyu and Wu, Jibin and Tan, Kay Chen},
  journal={arXiv preprint arXiv:2505.23433},
  year={2025}
}

@article{li2025choice,
  title={The choice of divergence: A neglected key to mitigating diversity collapse in reinforcement learning with verifiable reward},
  author={Li, Long and Zhou, Zhijian and Hao, Jiaran and Liu, Jason Klein and Miao, Yanting and Pang, Wei and Tan, Xiaoyu and Chu, Wei and Wang, Zhe and Pan, Shirui and others},
  journal={arXiv preprint arXiv:2509.07430},
  year={2025}
}

@article{zhou2024archer,
  title={Archer: Training language model agents via hierarchical multi-turn rl},
  author={Zhou, Yifei and Zanette, Andrea and Pan, Jiayi and Levine, Sergey and Kumar, Aviral},
  journal={arXiv preprint arXiv:2402.19446},
  year={2024}
}

@article{putta2024agent,
  title={Agent q: Advanced reasoning and learning for autonomous ai agents},
  author={Putta, Pranav and Mills, Edmund and Garg, Naman and Motwani, Sumeet and Finn, Chelsea and Garg, Divyansh and Rafailov, Rafael},
  journal={arXiv preprint arXiv:2408.07199},
  year={2024}
}

@article{bai2024digirl,
  title={Digirl: Training in-the-wild device-control agents with autonomous reinforcement learning},
  author={Bai, Hao and Zhou, Yifei and Cemri, Mert and Pan, Jiayi and Suhr, Alane and Levine, Sergey and Kumar, Aviral},
  journal={Advances in Neural Information Processing Systems},
  volume={37},
  pages={12461--12495},
  year={2024}
}

@article{feng2025towards,
  title={Towards efficient online tuning of vlm agents via counterfactual soft reinforcement learning},
  author={Feng, Lang and Tan, Weihao and Lyu, Zhiyi and Zheng, Longtao and Xu, Haiyang and Yan, Ming and Huang, Fei and An, Bo},
  journal={arXiv preprint arXiv:2505.03792},
  year={2025}
}

@article{chen2025reinforcement,
  title={Reinforcement learning for long-horizon interactive llm agents},
  author={Chen, Kevin and Cusumano-Towner, Marco and Huval, Brody and Petrenko, Aleksei and Hamburger, Jackson and Koltun, Vladlen and Kr{\"a}henb{\"u}hl, Philipp},
  journal={arXiv preprint arXiv:2502.01600},
  year={2025}
}

@article{wang2025ragen,
  title={Ragen: Understanding self-evolution in llm agents via multi-turn reinforcement learning},
  author={Wang, Zihan and Wang, Kangrui and Wang, Qineng and Zhang, Pingyue and Li, Linjie and Yang, Zhengyuan and Jin, Xing and Yu, Kefan and Nguyen, Minh Nhat and Liu, Licheng and others},
  journal={arXiv preprint arXiv:2504.20073},
  year={2025}
}

@article{feng2505group,
  title={Group-in-group policy optimization for llm agent training, 2025},
  author={Feng, Lang and Xue, Zhenghai and Liu, Tingcong and An, Bo},
  journal={arXiv preprint arXiv:2505.10978},
  year={2025}
}

@inproceedings{shridhar2020alfworld,
  title ={{ALFWorld: Aligning Text and Embodied
           Environments for Interactive Learning}},
  author={Mohit Shridhar and Xingdi Yuan and
          Marc-Alexandre C\^ot\'e and Yonatan Bisk and
          Adam Trischler and Matthew Hausknecht},
  booktitle = {Proceedings of the International Conference on Learning Representations (ICLR)},
  year = {2021},
  url = {https://arxiv.org/abs/2010.03768}
}

@misc{schrader2018gymsokoban,
  author       = {Schrader, Max-Philipp B.},
  title        = {Gym-Sokoban},
  year         = {2018},
  howpublished = {\url{https://github.com/mpSchrader/gym-sokoban}},
  note         = {GitHub repository}
}

@inproceedings{sheng2025hybridflow,
  title={Hybridflow: A flexible and efficient rlhf framework},
  author={Sheng, Guangming and Zhang, Chi and Ye, Zilingfeng and Wu, Xibin and Zhang, Wang and Zhang, Ru and Peng, Yanghua and Lin, Haibin and Wu, Chuan},
  booktitle={Proceedings of the Twentieth European Conference on Computer Systems},
  pages={1279--1297},
  year={2025}
}

@article{du2026asurvey,
  author       = {Shangheng Du and
                  Jiabao Zhao and
                  Jinxin Shi and
                  Zhentao Xie and
                  Xin Jiang and
                  Yanhong Bai and
                  Liang He},
  title        = {A Survey on the Optimization of Large Language Model-based Agents},
  journal      = {{ACM} Comput. Surv.},
  volume       = {58},
  number       = {9},
  pages        = {223:1--223:37},
  year         = {2026},
  url          = {https://doi.org/10.1145/3789261},
  doi          = {10.1145/3789261},
  bibsource    = {dblp computer science bibliography, https://dblp.org}
}

@article{zhang2026thelandscape,
  author       = {Guibin Zhang and
                  Hejia Geng and
                  Xiaohang Yu and
                  Zhenfei Yin and
                  Zaibin Zhang and
                  Zelin Tan and
                  Heng Zhou and
                  Zhong{-}Zhi Li and
                  Xiangyuan Xue and
                  Yijiang Li and
                  Yifan Zhou and
                  Yang Chen and
                  Chen Zhang and
                  Yutao Fan and
                  Zihu Wang and
                  Songtao Huang and
                  Francisco Piedrahita Velez and
                  Yue Liao and
                  Hongru Wang and
                  Mengyue Yang and
                  Heng Ji and
                  Jun Wang and
                  Shuicheng Yan and
                  Philip Torr and
                  Lei Bai},
  title        = {The Landscape of Agentic Reinforcement Learning for LLMs: {A} Survey},
  journal      = {Trans. Mach. Learn. Res.},
  volume       = {2026},
  year         = {2026},
  url          = {https://openreview.net/forum?id=RY19y2RI1O},
  bibsource    = {dblp computer science bibliography, https://dblp.org}
}

@inproceedings{orbert2024understand,
  author       = {Robert Kirk and
                  Ishita Mediratta and
                  Christoforos Nalmpantis and
                  Jelena Luketina and
                  Eric Hambro and
                  Edward Grefenstette and
                  Roberta Raileanu},
  title        = {Understanding the Effects of {RLHF} on {LLM} Generalisation and Diversity},
  booktitle    = {The Twelfth International Conference on Learning Representations,
                  {ICLR} 2024, Vienna, Austria, May 7-11, 2024},
  publisher    = {OpenReview.net},
  year         = {2024},
  url          = {https://openreview.net/forum?id=PXD3FAVHJT},
  bibsource    = {dblp computer science bibliography, https://dblp.org}
}

@inproceedings{yao2022webshop,
  title={Webshop: Towards scalable real-world web interaction with grounded language agents},
  author={Yao, Shunyu and Chen, Howard and Yang, John and Narasimhan, Karthik},
  booktitle={Advances in Neural Information Processing Systems (NeurIPS)},
  volume={35},
  pages={20744--20757},
  year={2022}
}

@article{cully2017quality,
  title={Quality and diversity optimization: A unifying modular framework},
  author={Cully, Antoine and Demiris, Yiannis},
  journal={IEEE Transactions on Evolutionary Computation},
  volume={22},
  number={2},
  pages={245--259},
  year={2017},
  publisher={IEEE}
}

@inproceedings{li2025longdiff,
  title={LongDiff: Training-Free Long Video Generation in One Go},
  author={Li, Zhuoling and Rahmani, Hossein and Ke, Qiuhong and Liu, Jun},
  booktitle={2025 IEEE/CVF Conference on Computer Vision and Pattern Recognition (CVPR)},
  pages={17789--17798},
  year={2025},
  organization={IEEE}
}

@inproceedings{li2026automatic,
  title={Automatic Method Illustration Generation for AI Scientific Papers via Drawing Middleware Creation, Evolution, and Orchestration},
  author={Li, Zhuoling and Zhang, Jiarui and Hu, Ping and Kuen, Jason and Gu, Jiuxiang and Rahmani, Hossein and Liu, Jun},
  booktitle={European Conference on Computer Vision},
  year={2026}
}

@inproceedings{li2026diffgraph,
  title={DiffGraph: An Automated Agent-driven Model Merging Framework for In-the-Wild Text-to-Image Generation},
  author={Li, Zhuoling and Rahmani, Hossein and Zhang, Jiarui and Xue, Yu and Mirmehdi, Majid and Kuen, Jason and Gu, Jiuxiang and Liu, Jun},
  booktitle={Proceedings of the IEEE/CVF Conference on Computer Vision and Pattern Recognition},
  year={2026}
}

@inproceedings{zhang2025mer,
  title={Mer-inspector: Assessing model extraction risks from an attack-agnostic perspective},
  author={Zhang, Xinwei and Hu, Haibo and Ye, Qingqing and Bai, Li and Zheng, Huadi},
  booktitle={Proceedings of the ACM on Web Conference 2025},
  pages={4300--4315},
  year={2025}
}

@article{zhang2026adversarial,
  title={On the Adversarial Robustness of Large Vision-Language Models under Visual Token Compression},
  author={Zhang, Xinwei and Liu, Hangcheng and Bai, Li and Wang, Hao and Ye, Qingqing and Zhang, Tianwei and Hu, Haibo},
  journal={arXiv preprint arXiv:2601.21531},
  year={2026}
}

@article{zhang2026semantics,
  title={Semantics-Aware Bilevel Co-Evolution: Towards Automated Multicomponent Algorithm Design},
  author={Zhang, Zhiyao and Wu, Shenghao and Wu, Xingyu and Tan, Kay Chen},
  journal={arXiv preprint arXiv:2606.29953},
  year={2026}
}

@article{zhang2024two,
  title={A two-phase kriging-assisted evolutionary algorithm for expensive constrained multiobjective optimization problems},
  author={Zhang, Zhiyao and Wang, Yong and Liu, Jiao and Sun, Guangyong and Tang, Ke},
  journal={IEEE Transactions on Systems, Man, and Cybernetics: Systems},
  volume={54},
  number={8},
  pages={4579--4591},
  year={2024},
  publisher={IEEE}
}

@inproceedings{wang2026learning,
  title={Learning while staying curious: Entropy-preserving supervised fine-tuning via adaptive self-distillation for large reasoning models},
  author={Wang, Hao and Gu, Hao and Piao, Hongming and Gong, Kaixiong and Ye, Yuxiao and Yue, Xiangyu and Han, Sirui and Guo, Yike and Wu, Dapeng},
  booktitle={Proceedings of the 64th Annual Meeting of the Association for Computational Linguistics (Volume 1: Long Papers)},
  pages={13567--13581},
  year={2026}
}

@inproceedings{wang2026deepmed,
  title={Deepmed: Building a medical deepresearch agent via multi-hop med-search data and turn-controlled agentic training \& inference},
  author={Wang, Zihan and Wang, Hao and Feng, Shi and Yang, Xiaocui and Wang, Daling and Zhang, Yiqun and Lin, Jinghao and Ji, Xiaozhong and Yang, Haihua},
  booktitle={Findings of the Association for Computational Linguistics: ACL 2026},
  pages={18160--18178},
  year={2026}
}

@article{yao2026slat,
  title={SLAT: Segment-Level Adaptive Trimming for Efficient CoT Reasoning},
  author={Yao, Jian and Luo, Xiongcai and Cheng, Ran and Tan, Kay Chen},
  journal={arXiv preprint arXiv:2605.30832},
  year={2026}
}

@article{yao2025var,
  title={Var-Math: Probing True Mathematical Reasoning in Llms via Symbolic Multi-Instance Benchmarks},
  author={Yao, Jian and Cheng, Ran and Tan, Kay Chen},
  journal={arXiv preprint arXiv:2507.12885},
  year={2025}
}

@inproceedings{
wang2026caseplay,
title={CasePlay: Self-Play Reinforcement Learning from Case Reports for Medical Reasoning},
author={Wang, Hao and Wang, Zihan and Wu, Kai and Yao, Jian and Ye, Yiwen and Luo, Yongcan and Wu, Dapeng and Chen, Jiale and Yang, Haihua},
booktitle={The Fortieth Annual Conference on Neural Information Processing Systems},
year={2026},
}

\clearpage
\appendix

\section{Derivation and Implementation of Trajectory-Level Diversity Credits}
\label{app:credit_derivation}

\paragraph{Pairwise diversity credit.}
For a sampled group of $G$ trajectory descriptors, define
\begin{equation}
\mathcal D_{\mathrm{pair}}(Z)
=
\frac{2}{G(G-1)}
\sum_{1\leq a<b\leq G}
d(\mathbf z_a,\mathbf z_b)
\end{equation}
and
\begin{equation}
c_i^{\mathrm{pair}}(Z)
=
\frac{1}{G-1}
\sum_{j\ne i}
d(\mathbf z_i,\mathbf z_j).
\end{equation}
Let
$S=\sum_{a<b}d(\mathbf z_a,\mathbf z_b)$ and
$s_i=\sum_{j\ne i}d(\mathbf z_i,\mathbf z_j)$.
Since
\begin{equation}
\mathcal D_{\mathrm{pair}}(Z_{-i})
=
\frac{2(S-s_i)}{(G-1)(G-2)},
\end{equation}
the exact leave-one-out change satisfies
\begin{equation}
\mathcal D_{\mathrm{pair}}(Z)
-
\mathcal D_{\mathrm{pair}}(Z_{-i})
=
\frac{2}{G-2}
\left[
c_i^{\mathrm{pair}}(Z)
-
\mathcal D_{\mathrm{pair}}(Z)
\right].
\label{eq:app_pairwise_relation}
\end{equation}
Moreover, every unordered pair contributes to two credits, so
\begin{equation}
\frac{1}{G}
\sum_{i=1}^{G}
c_i^{\mathrm{pair}}(Z)
=
\mathcal D_{\mathrm{pair}}(Z).
\end{equation}
Thus, within a fixed group, the exact leave-one-out change is a positive
affine transformation of $c_i^{\mathrm{pair}}$. The two therefore produce
identical values after group-wise standardization.

\paragraph{Determinantal diversity credit.}
We construct
\begin{equation}
K_{ij}
=
\exp\left(
-
\frac{\lVert\mathbf z_i-\mathbf z_j\rVert_2^2}{\sigma}
\right)
\end{equation}
and let $A=K+\eta I$. The implemented credit is
\begin{equation}
c_i^{\mathrm{det}}(Z)
=
\frac{\det(A)}
{\det(A_{-i,-i})}.
\label{eq:app_det_ratio}
\end{equation}
Equivalently,
\begin{equation}
\log c_i^{\mathrm{det}}
=
\log\det(A)
-
\log\det(A_{-i,-i}),
\end{equation}
so the determinant ratio is the exponential of the leave-one-out
log-determinant change. By the Schur-complement identity,
\begin{equation}
c_i^{\mathrm{det}}(Z)
=
A_{ii}
-
A_{i,-i}
A_{-i,-i}^{-1}
A_{-i,i}.
\end{equation}

\paragraph{Numerical implementation.}
The pairwise credit uses Euclidean distance. For the determinantal credit,
we use $\sigma=10$ and diagonal jitter $\eta=10^{-4}$. Raw
determinant-ratio credits are divided by their within-group maximum before
standardization; this common positive scaling does not affect the subsequent
group Z-score. If the regularized full determinant is at most $10^{-10}$,
we use the mean off-diagonal kernel dissimilarity $1-K_{ij}$ as a
numerically robust fallback. If an individual leave-one-out determinant is
at most $10^{-10}$ while the full determinant is valid, its raw credit is
set to one. Groups with zero credit variance receive zero normalized
diversity advantage.

\section{More Discussion on Related Work}
Given the recent surge of interest in RL for LLMs and agents, there is a substantial body of relevant work worth discussing. We provide a more detailed discussion of these prior studies here.

\paragraph{RL for LLMs and Agents.}
Recent LLMs have shown rapid progress in complex reasoning, with strong performance demonstrated by frontier reasoning models such as OpenAI-o1~\citep{openai-o1}, DeepSeek-R1~\citep{guo2025deepseek}, and Kimi-k1.5~\citep{team2025kimi}.
A representative line of work follows the R1-zero training paradigm introduced by DeepSeek-R1~\citep{guo2025deepseek}, which combines group-based policy optimization with rule-based rewards for final-answer correctness and formatting.
Subsequent studies have improved this paradigm along two main directions: improving data quality, curriculum construction, and rollout filtering~\citep{li2025limr,meng2023deepscaler,he2025deepmath,yu2503dapo,hu2025openreasonerzeroopensourceapproach,curatedthoughts}, and refining RL algorithms for more stable and efficient post-training.
The latter includes PPO-style and value-based methods for improved advantage estimation and long-chain credit assignment~\citep{zeng2025simplerl,kazemnejad2024vineppo,yuan2025s,yuan2025vapo}, GRPO-style methods that address instability, biased objectives, length or difficulty biases, and sample efficiency~\citep{yu2503dapo,liu2025understanding,zhang2025srpo,yao2026slat,yao2025var}, as well as REINFORCE-like methods that improve stability or scalability~\citep{team2025kimi,hu2025reinforce++,Chu_Huang_Zhang_Wei_Wang_2025}.
Relatedly, recent work has explored increasingly automated and modular generation pipelines across long-form video synthesis, agent-driven model merging, and scientific illustration generation~\citep{li2025longdiff,li2026diffgraph,li2026automatic}.

Beyond static reasoning tasks, RL has also been increasingly used to train LLM agents that interact
with external environments over multiple turns, including web navigation, application control, embodied
tasks, multi-hop information seeking, and other long-horizon interactive settings~\citep{wang2026deepmed,wang2026caseplay}.
Some methods improve agent learning through hierarchical training or guided search, often targeting web-based tasks such as WebShop~\citep{zhou2024archer,putta2024agent}.
Another line studies online RL for interactive agents in device-control or vision-language environments, where exploration in large action spaces is a central challenge~\citep{bai2024digirl,feng2025towards}.
More recent work adapts group-based or trajectory-level RL to long-horizon agent training, focusing on full-trajectory optimization, scalable multi-turn learning, or finer-grained credit assignment~\citep{chen2025reinforcement,wang2025ragen,feng2505group}.
These works primarily target training stability, exploration efficiency, and credit assignment in interactive environments.

\section{Additional Experimental Details}
\label{app:exp_details}

\subsection{Environment Details}
\label{app:env_details}

\paragraph{ALFWorld.}
ALFWorld~\citep{shridhar2020alfworld} is a text-based interactive environment aligned with embodied household tasks.
Each episode provides a natural-language goal and textual observations of the current environment.
At each step, the agent outputs a textual action, such as navigation, object manipulation, or environment inspection.
An episode is considered successful if the agent completes the specified household task within the step budget.
In our experiments, a trajectory consists of the full sequence of observations, reasoning tokens, actions, and environment feedback:
\begin{equation}
    \tau = (o_1, a_1, o_2, a_2, \dots, o_T, a_T).
\end{equation}
The task reward is defined as
\begin{equation}
    r^{\mathrm{task}} =
    \mathbf{1}[\text{task success}].
\end{equation}

\paragraph{Sokoban.} 
Sokoban ~\citep{schrader2018gymsokoban} is a grid-based box-pushing puzzle game.
The agent moves on a grid and must push boxes onto target cells.
The agent can push boxes but cannot pull them, making the environment sensitive to irreversible action sequences and long-horizon planning.
An episode is successful (and receive positive reward) if all boxes are placed on target cells within the step budget.
A trajectory consists of the sequence of grid states and actions:
\begin{equation}
    \tau = (s_1, a_1, s_2, a_2, \dots, s_T, a_T).
\end{equation}
We use Sokoban to evaluate whether explicit diversity can capture planning-level behavioral variation rather than tool-use variation.

\subsection{Training Implementation}
\label{app:training_details}

We implement our method on top of GiGPO~\citep{feng2505group} using the verl training framework~\citep{sheng2025hybridflow}.
For each input instance, we sample a group of $G$ trajectories from the current policy.
We compute the task reward $r_i^{\mathrm{task}}$ and descriptor vector $\mathbf{z}_i$ for each trajectory.
Given the descriptor matrix $Z$, we compute the diversity credit $c_i(Z)$
and standardize it within trajectories sampled for the same input:
\begin{equation}
    \hat c_i
    =
    \frac{c_i-\bar c}
    {\max\{\operatorname{std}(c),10^{-6}\}}.
\end{equation}
For both ALFWorld and Sokoban, the standardized credit is added to the
GiGPO advantage:
\begin{equation}
    \widetilde A_{i,t}
    =
    A_{i,t}^{\mathrm{GiGPO}}
    +
    \lambda \hat c_i.
\end{equation}
The trajectory-level signal is broadcast over the valid response tokens
associated with trajectory $i$.
The environment-level setup is summarized in Table~\ref{tab:env_setup}, and
the shared optimization and evaluation settings are reported in
Table~\ref{tab:hyperparams}.

\begin{table}[t]
\centering
\caption{Environment-specific experimental setup.}
\label{tab:env_setup}
\begin{tabular}{lcc}
\toprule
\textbf{Setting} & \textbf{ALFWorld} & \textbf{Sokoban} \\
\midrule
Base model & Qwen2.5-7B-Instruct & Qwen3-VL-4B-Instruct \\
Input modality & Text & Vision-language \\
Environment & AlfredTWEnv & Sokoban \\
Environment observation & Text & RGB image \\
Max environment steps & 50 & 15 \\
Training data size & 16 & 32 \\
Total epochs & 150 & 200 \\
Advantage normalization mode & mean\_std\_norm & mean\_norm \\
Max prompt length & 2048 & 1024 \\
PPO mini-batch size & 256 & 64 \\
PPO micro-batch size per GPU & 16 & 8 \\
Validation frequency (steps) & 5 & 10 \\
\bottomrule
\end{tabular}
\end{table}

\begin{table}[t]
\centering
\caption{Shared training and evaluation hyperparameters.}
\label{tab:hyperparams}
\begin{tabular}{lc}
\toprule
\textbf{Hyperparameter} & \textbf{Value} \\
\midrule
Training framework & verl \\
RL backbone & GiGPO \\
Group size $G$ & 8 \\
Max response length & 512 \\
Learning rate & $1\times 10^{-6}$ \\
KL loss type & low-variance KL \\
KL loss coefficient & 0.01 \\
Discount factor $\gamma$ & 0.95 \\
GiGPO step-advantage weight $\omega$ & 1.0 \\
Diversity integration mode & advantage \\
Diversity-credit normalization & group mean/std \\
Pairwise distance & Euclidean \\
DPP kernel bandwidth $\sigma$ & 10 \\
DPP diagonal jitter $\eta$ & $10^{-4}$ \\
Diversity-credit standard-deviation floor & $10^{-6}$ \\
Invalid action penalty coefficient & 0.1 \\
Validation temperature & 0.4 \\
Standard evaluation presence penalty & 0.0 \\
Frequency penalty & 0.0 \\
Standard repetition penalty & 1.0 \\
Stochastic validation decoding & yes \\
Evaluation trajectories per input $N_{\mathrm{eval}}$ & 8 \\
Number of GPUs & 8 \\
Number of nodes & 1 \\
Tensor parallel size & 1 \\
GPU memory utilization & 0.6 \\
\bottomrule
\end{tabular}
\end{table}

\subsection{Reproducibility}
\label{app:reproducibility_details}

\paragraph{Runs and randomness.}
Each reported model configuration is obtained from one training run.
The training scripts set the environment seed to $0$ for both ALFWorld and
Sokoban.
Evaluation uses stochastic decoding with temperature $0.4$.
For the main ALFWorld comparison, every checkpoint is evaluated on the same
128 prompts with eight rollouts per prompt, yielding 1,024 trajectories per
checkpoint.
The GiGPO+PP and GiGPO+RP baselines reuse the selected GiGPO checkpoint and
change only the vLLM presence penalty or repetition penalty, respectively.
They receive no additional training and are not used for checkpoint selection.
Diversity metrics are computed within prompt groups over successful
trajectories.

\paragraph{Compute and software.}
All reported training runs use one node with eight GPUs and tensor parallel
size one.
The experiments use eight NVIDIA H20 GPUs with 144\,GB memory per GPU.
Training is implemented in the PyTorch-based verl framework, with Ray for
environment workers and vLLM for rollout generation; the environment uses
vLLM 0.17, and pins Transformers 4.51.1.

\section{Trajectory Descriptor Details}
\label{app:bd_details}

This section describes the task-specific trajectory descriptors used in our experiments.
Each descriptor maps a trajectory $\tau$ to a vector $\mathbf{z}(\tau)\in\mathbb{R}^m$.
All descriptor values are computed after trajectory generation and are used
as policy-optimization signals during RL training.

\subsection{ALFWorld Descriptors}

 \paragraph{Action-type descriptor.}
 The action-type descriptor captures the distribution of high-level action
 categories used by the agent across a trajectory. We parse each environment
 action into one of $m{=}15$ canonical ALFWorld action types: navigation
 (\textit{go to}), container manipulation (\textit{open}, \textit{close}),
 object manipulation (\textit{pick up}, \textit{put}, \textit{take},
 \textit{slice}), state-changing operations (\textit{clean}, \textit{heat},
 \textit{cool}, \textit{toggle}, \textit{use}), and inspection
 (\textit{examine}, \textit{look}, \textit{inventory}). Action-type matching
 is performed by string-prefix lookup on the parsed action text. Let
 $\mathcal{A}_{\mathrm{type}}=\{c_1,\dots,c_m\}$ denote the action-category
 set. For trajectory $\tau=\{a_1,\dots,a_T\}$, the descriptor is the
 normalized count vector
 \begin{equation}
     \phi^{\mathrm{action}}_k(\tau)
     \;=\;
     \frac{1}{T}
     \sum_{t=1}^{T}
     \mathbf{1}\!\left[\mathrm{type}(a_t)=c_k\right],
     \label{eq:bd-action-type}
 \end{equation}
 for $k=1,\dots,m$,
 yielding $\phi^{\mathrm{action}}(\tau)\in\Delta^{m-1}\subset\mathbb{R}^{15}$.
 This descriptor encourages diversity in how the agent structures its
 interaction process, e.g., whether it relies more on navigation,
 inspection, or manipulation, and is invariant to the specific objects or
 locations involved.

 \paragraph{Location descriptor.}
 The location descriptor captures the distribution of navigation destinations
 visited by the agent during a trajectory. For an ALFWorld input $x$, let
 $\mathcal{L}(x)=\{\ell_1,\dots,\ell_n\}$ denote the locations available in
 the corresponding environment. We parse each navigation action of the form
 \textit{go to} $\ell_k$ and define the normalized location-frequency vector
 \begin{equation}
     \phi^{\mathrm{location}}_k(\tau)
     \;=\;
     \frac{1}{Z(\tau)}
     \sum_{t=1}^{T}
     \mathbf{1}\!\left[
     \substack{
         \mathrm{type}(a_t)=\textit{go\,to}\\
         \mathrm{location}(a_t)=\ell_k
     }
     \right],
     \label{eq:bd-location-freq}
 \end{equation}
 for $k=1,\dots,n$, where $Z(\tau)$ is the total number of parsed navigation actions.
 This yields
 $\phi^{\mathrm{location}}(\tau)\in\Delta^{n-1}\subset
 \mathbb{R}^{|\mathcal{L}(x)|}$.
 The descriptor encourages diversity in route choice and location visitation
 patterns while remaining invariant to non-navigation action composition.

\subsection{Sokoban Descriptors}

 For Sokoban we instantiate two complementary trajectory-level behavior
 descriptors that operate on the agent's emitted move sequence:
 a \emph{direction-transition} descriptor that summarizes transitions
 between consecutive move directions, and a higher-dimensional \emph{spatial-coverage}
 descriptor that captures \emph{where} on the grid the agent traverses.
 Both descriptors are extracted from the action stream by parsing
 action tags inside the agent's
 response and are $\ell_1$-normalized to lie on the probability simplex.

 \paragraph{Direction-transition descriptor.}
 The direction-transition descriptor captures the distribution of
 consecutive low-level move-direction pairs used by the agent. Let
 $\mathcal{D}=\{d_1,d_2,d_3,d_4\}=\{\textit{up},\textit{down},\textit{left},
 \textit{right}\}$ denote the four cardinal moves, and let $T$ be the
 number of valid move actions in trajectory $\tau$. For each ordered pair
 $(d_p,d_q)\in\mathcal{D}\times\mathcal{D}$, we define
 \begin{equation}
     \phi^{\mathrm{dir}}_{p,q}(\tau)
     \;=\;
     \frac{1}{T-1}
     \sum_{t=1}^{T-1}
     \mathbf{1}\!\left[
     \substack{
         \mathrm{dir}(a_t)=d_p\\
         \mathrm{dir}(a_{t+1})=d_q
     }
     \right],
     \label{eq:bd-direction-transition}
 \end{equation}
 yielding a normalized transition histogram
 $\phi^{\mathrm{dir}}(\tau)\in\Delta^{15}\subset\mathbb{R}^{16}$.
 This descriptor is invariant to the absolute position on the board and
 encourages diversity in movement patterns, including straight runs,
 reversals, and changes between horizontal and vertical motion.

 \paragraph{Spatial-coverage descriptor.}
 The spatial-coverage descriptor captures \emph{where} the agent moves
 on the grid by simulating its trajectory in a board-local coordinate
 frame. We instantiate a $H{\times}W$ relative grid (we use
 $H{=}W{=}7$ in our experiments) centered at the agent's start cell,
 and roll out the action sequence with a clipping policy that keeps
 the simulated agent inside the grid. Let
 $\mathcal{G}=\{g_1,\dots,g_m\}$ with $m{=}HW$ enumerate the grid
 cells, and let $c_t\in\mathcal{G}$ be the simulated cell at step $t$
 (with $c_0$ the start cell). The descriptor is the normalized
 visitation count
 \begin{equation}
     \phi^{\mathrm{cov}}_k(\tau)
     \;=\;
     \frac{1}{Z(\tau)}
     \sum_{t=0}^{T}
     \mathbf{1}\!\left[c_t = g_k\right],
     \qquad k=1,\dots,m,
     \label{eq:bd-spatial-coverage}
 \end{equation}
 where $Z(\tau)=T+1$ normalizes the histogram to the simplex,
 $\phi^{\mathrm{cov}}(\tau)\in\Delta^{m-1}\subset\mathbb{R}^{49}$.
 Because the simulator anchors the origin at the agent's start cell
 rather than the absolute board frame, two trajectories on the same
 puzzle that visit the same physical cells yield identical descriptors,
 while trajectories taking distinct routes are pulled apart in
 $\phi^{\mathrm{cov}}$-space. This descriptor encourages diversity in
 spatial exploration and route choice, and is complementary to
 $\phi^{\mathrm{dir}}$: two policies can share a similar direction
 distribution yet differ sharply in which grid cells they cover.

\subsection{Descriptor Summary}
\begin{table}[t]
\centering
\caption{Summary of trajectory descriptors used in our experiments.}
\label{tab:descriptor_summary}
\begin{tabularx}{\linewidth}{@{}ll>{\raggedright\arraybackslash}X>{\raggedright\arraybackslash}X@{}}
\toprule
\textbf{Environment} & \textbf{Descriptor} & \textbf{What it captures} & \textbf{Vector type} \\
 \midrule
 ALFWorld & action\_type          & High-level action          & normalized histogram
($\mathbb{R}^{15}$) \\
 ALFWorld & location              & Navigation destination usage           & normalized histogram
($\mathbb{R}^{|\mathcal{L}(x)|}$) \\
 Sokoban  & direction\_transition & Consecutive direction pairs            & normalized histogram
($\mathbb{R}^{16}$)  \\
 Sokoban  & spatial\_coverage     & Visited grid-cell distribution         & normalized 2D occupancy
($\mathbb{R}^{7{\times}7}$) \\
 \bottomrule
\end{tabularx}
\end{table}

Table~\ref{tab:descriptor_summary} summarizes how the four descriptors capture
complementary behavioral variation in action composition, location visitation,
direction transitions, and spatial coverage.

\section{Determinantal Diversity Evaluation metric.}
\label{app:alfworld_determinantal_eval}
For the determinantal comparison, we report a size-normalized
evaluation score derived from the log-determinant functional.
Let $Z_{\mathrm{succ}}$ contain the descriptors of the $n$ successful
trajectories in an evaluation group, and let $K_{\mathrm{succ}}$
be their RBF Gram matrix.
For $n\geq2$, we define
\begin{equation}
D^{\mathrm{dpp}}(Z_{\mathrm{succ}})
=
\exp\!\left(
\frac{1}{n}\log\det(K_{\mathrm{succ}}+\eta I)
\right)
=
\det(K_{\mathrm{succ}}+\eta I)^{1/n},
\label{eq:eval_dpp_root}
\end{equation}
where
$K_{ij}=\exp(-\lVert z_i-z_j\rVert_2^2/10)$
and $\eta=10^{-4}$.
Here, $n$ is the number of successful trajectories, not the
descriptor dimension.
Groups with fewer than two successes are assigned zero.
The transformation is applied before aggregation and is used only
for evaluation; it does not change the training-credit definition.

\section{Location-Diversity Case Study}
\label{app:location_case_study}

\paragraph{Alternative Navigation Routes.}
\begin{figure}[t]
\centering
\includegraphics[width=\textwidth]{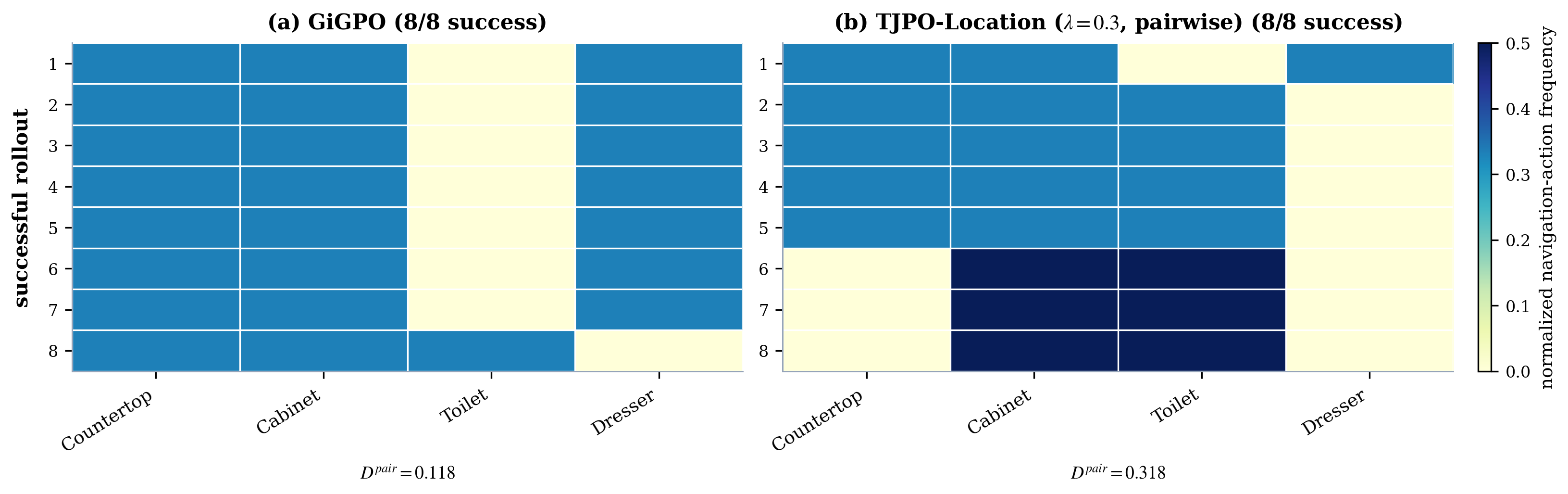}
\caption{
Location-descriptor view of the main-paper route case study for the prompt
``put a spraybottle in cabinet.''
Each row is one successful rollout, each column is a location activated by at
least one rollout, and color denotes the normalized frequency of navigation
actions targeting that location.
Rows are ordered by their projection onto the first principal component of
the pooled GiGPO and TJPO descriptor vectors, and both panels use a shared
color scale.
}
\label{fig:alfworld_location_case}
\end{figure}

Figure~\ref{fig:alfworld_location_case} provides the descriptor-level view of
the same eight-rollout comparison visualized as route graphs in the main
paper.
The GiGPO rows are nearly identical: seven rollouts allocate equal navigation
frequency to Countertop, Dresser, and Cabinet, reflecting the dominant
Countertop--Dresser--Cabinet route.
In contrast, the TJPO descriptors separate into multiple patterns, including
rollouts that navigate directly through the Toilet and rollouts that retain
the Dresser alternative.
This greater between-rollout heterogeneity raises
$D_{\mathrm{loc}}^{\mathrm{pair}}$ from $0.118$ to $0.318$.
Together, the route graph and heatmap show that the increase comes from
distinct successful navigation strategies rather than a change in task
success; both methods solve all eight rollouts.

\begin{figure}[t]
\centering
\includegraphics[width=\textwidth]{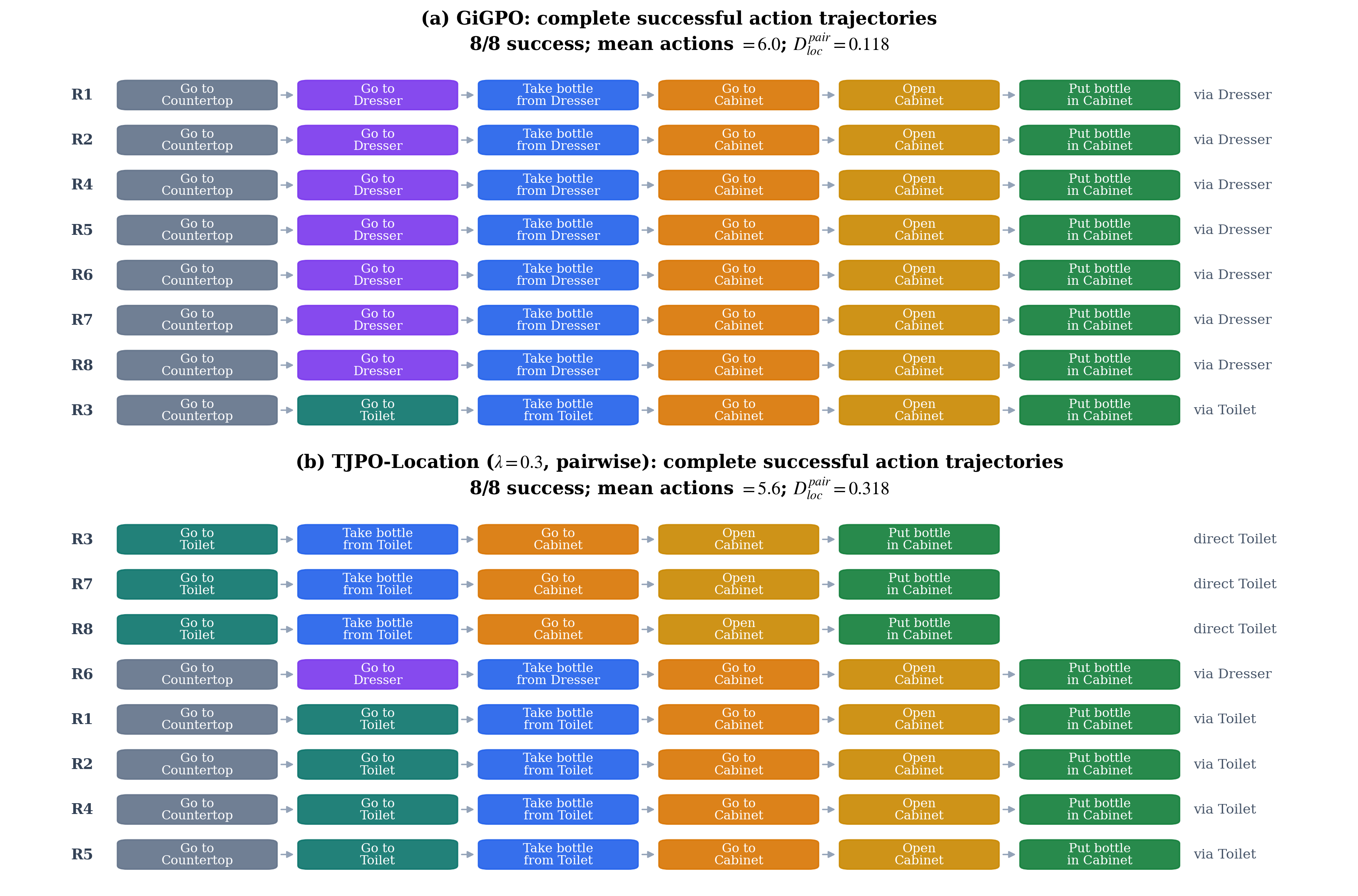}
\caption{
Complete successful environment-action trajectories for the same
``put a spraybottle in cabinet'' case.
All eight sampled rollouts are shown for each method; no trajectories are
omitted or selected.
Rollouts are grouped by navigation strategy while retaining their original
rollout identifiers.
Object instance indices are omitted for readability, but every environment
action is shown.
Colors distinguish navigation destinations and task-completion operations.
}
\label{fig:alfworld_location_full_trajectories}
\end{figure}

Figure~\ref{fig:alfworld_location_full_trajectories} exposes where the
trajectory diversity arises.
For GiGPO, seven of eight rollouts execute the same six-action sequence through
the Countertop and Dresser; the remaining rollout replaces the Dresser with
the Toilet but otherwise follows the same plan, yielding two unique route
signatures.
TJPO instead produces three successful strategy groups: three five-action
rollouts navigate directly to the Toilet, four six-action rollouts inspect the
Countertop before using the Toilet, and one rollout retains the Dresser route.

The variation is localized to the search and navigation prefix.
After a spray bottle is found, all rollouts use the same task-completion
structure: take the bottle, navigate to the Cabinet, open it, and place the
bottle inside.
Thus, TJPO changes which successful route is selected without disrupting the
reliable manipulation sequence.
The three direct-Toilet trajectories are also one action shorter than the
GiGPO trajectories, and none of the TJPO rollouts revisits a location.
All 16 displayed trajectories contain zero invalid actions and zero location
revisits.
The increase in $D_{\mathrm{loc}}^{\mathrm{pair}}$ therefore reflects
alternative efficient routes rather than longer or repetitive exploration.

\begin{figure}[p]
\centering
\includegraphics[width=\textwidth]{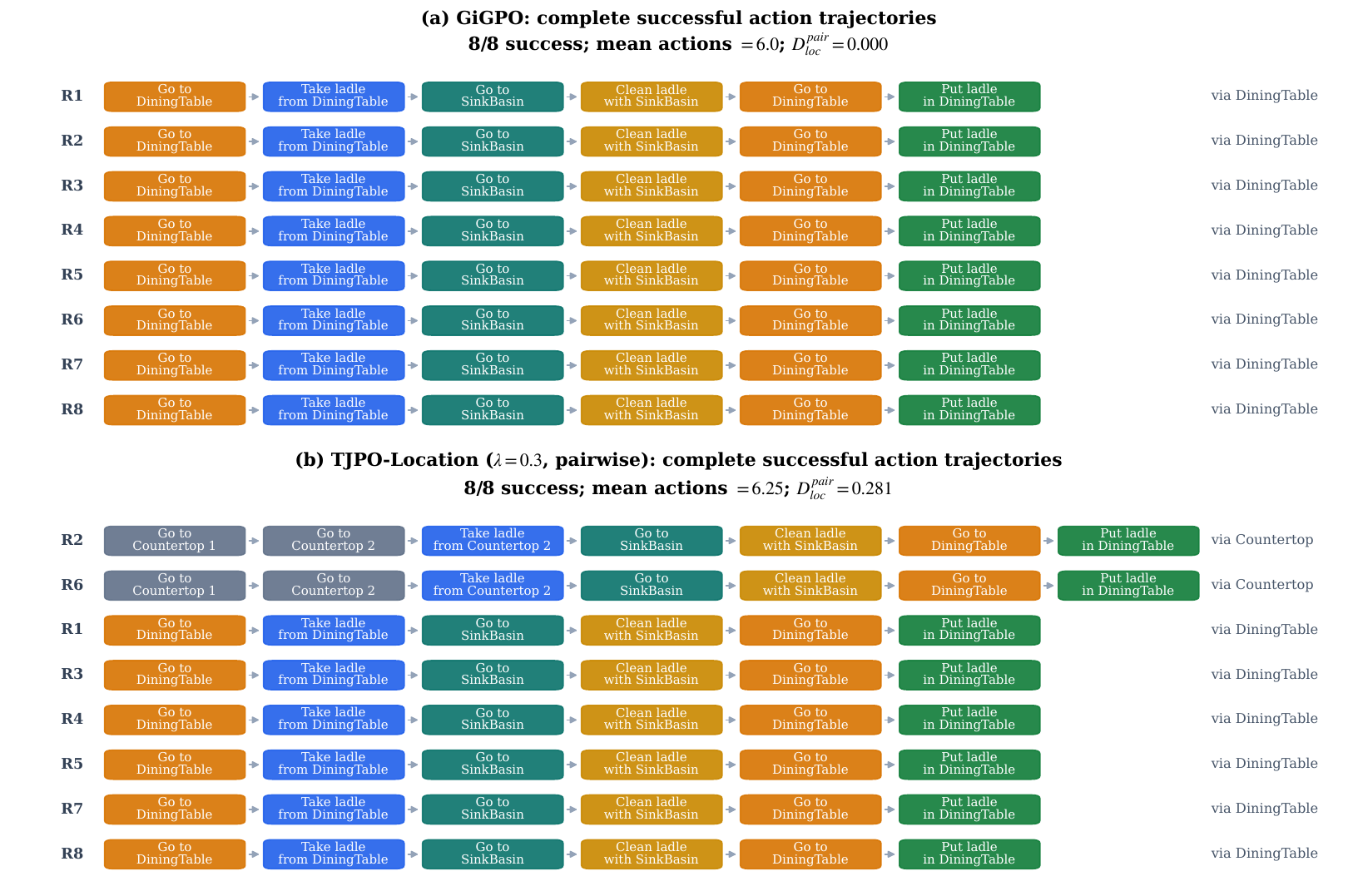}
\caption{
Complete successful trajectories for
``clean some ladle and put it in diningtable.''
The top panel shows GiGPO and the bottom panel shows location-pairwise TJPO
with $\lambda=0.3$.
All eight rollouts are retained for each method, with their original
rollout identifiers preserved.
Object instance indices and singleton-location indices are omitted
for readability, while Countertop~1 and Countertop~2 are distinguished.
Rows are grouped by navigation sequence; both panels use the same
action-position scale, and no actions are omitted.
}
\label{fig:additional_ladle_cleaning}
\end{figure}

\paragraph{Alternative Retrieval Locations before Cleaning.}
Figure~\ref{fig:additional_ladle_cleaning} provides an additional
location-diversity example for ``clean some ladle and put it in diningtable.''
GiGPO follows the same six-action sequence in all eight rollouts:
retrieve a ladle from the DiningTable, clean it at the SinkBasin,
and return it to the DiningTable.
TJPO follows this branch in six rollouts, while two rollouts retrieve
a different ladle from a Countertop before completing the same cleaning
and placement operations.
Both methods succeed in all eight rollouts.
The location-diversity score increases from $0$ to $0.281$,
while the mean action count changes from $6.00$ to $6.25$.
This example illustrates variation in retrieval locations while preserving
the required processing and final placement operations.

\section{Practical Benefits of Behavioral Diversity}
\label{app:known_location_protocol}

\paragraph{Evaluation suite.}
Table~\ref{tab:location_access} reports results on a structurally verified
subset of ALFWorld test games, containing 15 tasks from 14 scenes:
nine Pick \& Place, four Clean \& Place, and two Cool \& Place tasks.
Each task admits successful solutions through two independent source
location categories. We retain both intervention directions, yielding
30 cases for each of the User Constraints and Disruption conditions.

\paragraph{Task construction.}
Tasks are selected using PDDL, task metadata, and native environment
replay, without using model scores.
The two solution branches must use different target object instances.
We exclude tasks with shared target objects, or with a forbidden location
sharing its symbolic position with an unrestricted named receptacle.
Source categories must not overlap the final goal category or required
processing tools.
Each branch must avoid the other source category, and the lengths of
their verified plans must differ by at most two actions.
These are feasible witness lengths, not claimed shortest-path lengths.

For each source category $X$, the intervention covers all initially
navigable named locations assigned to that category.
We verify successful execution using $X$, avoiding $X$ in the original
environment, avoiding $X$ under both intervention conditions, and
completing an alternative plan after one rejected access to $X$.
Every successful replay must confirm target pickup and final placement
within 50 steps.
Both directions must pass these checks; all admitted tasks and directions
are retained in the evaluation.

\paragraph{Evaluation conditions.}
\emph{Standard} uses the original task without additional instructions
or changes to the environment.

The \emph{User Constraints} condition adds an instruction to avoid
locations of type $X$, explicitly lists the corresponding locations,
and asks the agent to complete the original task.
The instruction is repeated in each observation.
This condition does not modify admissible actions or environment dynamics,
so the agent can violate the requirement.

The \emph{Disruption} condition makes the same predefined location set
unavailable to both models from the start of the episode, without
initially informing the agent.
An attempted navigation or interaction involving this set is rejected
and replaced by \textit{look}, preserving the manipulation state while
consuming one interaction step.
The first rejection reveals the disruption.
Subsequent observations and admissible actions reflect the persistent
unavailability of these locations.
Neither intervention supplies an alternative location or action plan.

\paragraph{Models and sampling.}
We compare GiGPO with location-pairwise TJPO and $\lambda=0.5$.
No additional training is performed for these interventions.
Each case receives eight stochastic rollouts at temperature $0.4$
with a maximum of 50 interaction steps.
The environment seed is fixed to zero.
For each model, the evaluation therefore contains 120 Standard,
240 User Constraints, and 240 Disruption trajectories.

\paragraph{Metrics and aggregation.}
Task success is determined by the native environment.
Standard Success is the proportion of successful episodes without
intervention.
User Constraints Success requires both task completion and no
environment-confirmed visit to the forbidden location set.
Disruption Success is the proportion of successful episodes under
location unavailability, including episodes that never attempt to access
a forbidden location.

\section{Broader Impact}
\label{app:broader_impact}

This work studies how to explicitly define and optimize trajectory diversity in RL-based post-training for LLM agents.
A potential positive impact is improved controllability and robustness: by encouraging multiple high-quality behavioral modes, agents may become less dependent on a single brittle solution path and better adapt to changes in tools, interfaces, or environments.
The use of user-specified, task-specific descriptors also makes the diversity objective more interpretable, allowing practitioners to specify which forms of behavioral variation are desirable for a given application.
Robustness is also an important concern for multimodal models, where architectural or efficiency-oriented
interventions can interact nontrivially with adversarial behavior~\citep{zhang2026adversarial,zhang2025mer}.

At the same time, explicit diversity should be used together with appropriate safety constraints and task validation.
In applications involving external tools or real-world operations, practitioners should ensure that diversified behaviors remain safe, reliable, and aligned with intended use.

\end{document}